\documentclass{article}
\usepackage[utf8]{inputenc}
\usepackage{cite}
\usepackage{amsmath}
\usepackage{algorithmic}
\usepackage{array}
\usepackage[caption=false,font=footnotesize]{subfig}
\usepackage{fixltx2e}
\usepackage{url}
\usepackage{graphicx}
\usepackage{pgfplots}
\pgfplotsset{compat=1.8}
\usepackage[margin=1in]{geometry}
\usepackage{stmaryrd}
\usepackage{amsfonts}
\usepackage{amssymb}
\usepackage{array}
\usepackage{algorithm}
\usepackage{algorithmic}
\usepackage{mathrsfs}
\usepackage{relsize}
\usepackage{cite}
\usepackage{enumerate}
\usepackage{url}
\usepackage{lipsum}
\usepackage{booktabs} 
\usepackage{multirow}
\usepackage{dsfont}
\usepackage{latexsym}
\usepackage{hhline}
\usepackage{color}
\usepackage{textcomp}
\usepackage{hyperref}
\usepackage{bbding} 

\newcommand{\norm}[1]{\left\lVert \, #1 \, \right\rVert}
\newcommand{\abs}[1]{\left\lvert \, #1 \, \right\rvert}
\newcommand{\KOp}[1]{\llbracket #1 \rrbracket} 

\newcommand{\vect}[1]{\mathbf{#1}}

\newcommand{\vc}{\mathbf{c}}

\newcommand{\vx}{\mathbf{x}}
\newcommand{\vy}{\mathbf{y}}
\newcommand{\vz}{\mathbf{z}}

\newcommand{\vmu}{\boldsymbol{\mu}}
\newcommand{\vdelta}{\boldsymbol{\delta}}

\newcommand{\matr}[1]{\mathbf{#1}}
\newcommand{\mA}{\matr{A}}
\newcommand{\mB}{\matr{B}}
\newcommand{\mC}{\matr{C}}
\newcommand{\mD}{\matr{D}}

\newcommand{\mH}{\matr{H}}
\newcommand{\mI}{\matr{I}}

\newcommand{\mT}{\matr{T}}

\newcommand{\mX}{\matr{X}}

\newcommand{\mZ}{\matr{Z}}
\newcommand{\mM}{\matr{M}}

\newcommand{\mDelta}{\matr{\Delta}}

\newcommand{\R}{\mathbb{R}}
\newcommand{\N}{\mathbb{N}}

\makeatletter
\newcommand{\argmin}{\mathop{\text{~argmin~}}}

\DeclareMathOperator{\prox}{prox}  
\DeclareMathOperator{\vecn}{vec}

\newcommand{\tens}[1]{\boldsymbol{\mathcal{#1}}}

\newcommand{\tT}{\tens{T}}
\newcommand{\tX}{\tens{X}}
\newcommand{\tY}{\tens{Y}}

\newcommand{\tN}{\tens{N}}

\begin{document}
\title{A Flexible Optimization Framework for Regularized Matrix-Tensor Factorizations with
Linear Couplings}

\author{Carla~Schenker\thanks{C. Schenker is with the Simula Metropolitan Center for Digital Engineering and Oslo Metropolitan University, Oslo, Norway (e-mail: carla@simula.no).},
        Jeremy~E.~Cohen\thanks{J. E. Cohen is with CNRS, Université de Rennes, Inria, CNRS, IRISA.}~
        and~Evrim~Acar\thanks{E. Acar is with the Simula Metropolitan Center for Digital Engineering, Oslo, Norway.}
}
\date{}
\maketitle

\begin{abstract}
Coupled matrix and tensor factorizations (CMTF) are frequently used to jointly analyze data from multiple sources, also called data fusion. However, different characteristics of datasets stemming from multiple sources pose many challenges in data fusion and require to employ various regularizations, constraints, loss functions and different types of coupling structures between datasets. In this paper, we propose a flexible algorithmic framework for coupled matrix and tensor factorizations which utilizes Alternating Optimization (AO) and the Alternating Direction Method of Multipliers (ADMM). The framework facilitates the use of a variety of constraints, loss functions and couplings with linear transformations in a seamless way. Numerical experiments on simulated and real datasets demonstrate that the proposed approach is accurate, and computationally efficient with comparable or better performance than available CMTF methods for Frobenius norm loss, while being more flexible. Using Kullback-Leibler divergence on count data, we demonstrate that the algorithm yields accurate results also for other loss functions.
\end{abstract}



\section{Introduction}
In many areas of science, various sensing technologies are used to obtain information about a single system of interest. Often, none of the datasets alone contains a complete view of the system, but the data measured from different modalities can complement each other.
For instance, brain activity patterns 
can be captured using both electroencephalography (EEG) and functional magnetic resonance imaging (fMRI) signals, which have complementary temporal and spatial resolutions. Similarly, in metabolomics, multiple analytical techniques such as LC-MS (Liquid Chromatography - Mass Spectrometry) and NMR (Nuclear Magnetic Resonance) spectroscopy are used to measure chemical compounds in biological samples, providing a more complete picture of underlying biological processes.
Joint analysis of datasets from multiple sources, also referred to as data fusion (or multi-modal data mining), exploits these complementary measurements, and allows for better interpretability and, potentially, more accurate recovery of patterns characterizing the underlying phenomena. Nevertheless, data fusion poses many challenges, and there is an emerging need for data fusion methods that can take into account different characteristics of data from multiple sources in many disciplines \cite{MeSm10, Acar2015Data, SoBaLa15, LaAdJu15}.

Data from multiple sources can often be represented in the form of matrices and higher-order tensors. Coupled matrix and tensor factorizations (CMTF) are an effective approach for joint analysis of such datasets in many domains including social network analysis \cite{LiSuCaKo09, ZhCaZhXiYa10, ErAcCe13, ArRiFa17}, neuroscience \cite{PaMiSiFa14, EyHuDe17,ChDaEsKoTh18, AcShLe19,Rivet2015Multimodal}, and chemometrics \cite{Acar2015Data,WuAcKo18}. In such coupled factorizations, each dataset is modelled by a low-rank approximation. One of the most popular tensor factorization methods, Canonical Polyadic Decomposition (CPD) (also known as CANDECOMP/PARAFAC (CP))~\cite{Hi27a, Ha70, CaCh70} models a tensor $\tT_i$ of order $D_i\geq 2$ as the sum of $R_i$ (usually a small number) rank-one components,
\begin{equation}\label{eq:CP1}\small{
        \tT_i \approx \sum_{r=1}^{R_i} \mC_{i,1}(:,r)  \circ \mC_{i,2}(:,r)
         \circ ... \circ \mC_{i,D_i}(:,r)
         =: \KOp{\mC_{i,d}}_{d=1}^{D_i},}
\end{equation}
where $\mC_{i,d}(:,r)$ is the $r$-th column of factor matrix $\mC_{i,d}$, and \(\circ\) denotes the vector outer product. Those components may reveal patterns of interest in the data. Coupled factorization problems are often formulated by extracting the same latent factors from the coupled mode \cite{ZhCaZhXiYa10, AcKoDu11b, ErAcCe13, PaMiSiFa14, ArRiFa17}, as in the following formulation, where a third-order tensor $\tT_1$ is coupled with a matrix $\mT_2$ in the first mode:
\begin{equation}\small{
\begin{aligned}
    & \underset{\left\lbrace\mC_{i,d}\right\rbrace_{\underset{d\leq D_i}{i=1,2}}}{\min}
    \hspace*{-0em} & &
    \norm{\tT_1-  \KOp{\mC_{1,1}, \mC_{1,2}, \mC_{1,3}}}_F^2
  +  \norm{\smash{\mT_2-  \mC_{2,1}\mC_{2,2}^T}}_F^2 \\
 & \ \ \ \ \ \  \text{s.t. } & &
\mC_{1,1}=\mC_{2,1}
\end{aligned}}\label{eq:cmtf}
\end{equation}

However, factors corresponding to the coupled mode are not necessarily equal in datasets from different modalities. Therefore, in this paper, we focus on more general linear coupling relationships. In particular, we cover the cases, where not all factors are shared between tensors~\cite{ChDaEsKoTh18,YoKiKaCh10,Acar2015Data,DeWiBeAn11, KhLeKa16,YoCiYa12,SiErCeAc13}. For the time being, we assume that the number of shared components is known or estimated beforehand.
Moreover, linear transformations are also used to couple factors that are sampled from a continuous phenomenon with different sampling rates \cite{CabralFarias2016Exploring} or aggregation intervals \cite{AlKaSi20}.
Using such linear couplings, different resolutions of EEG and fMRI signals have been previously incorporated while jointly analyzing data from these two modalities \cite{EyHuDe17,ChDaEsKoTh18}. Linear couplings for spatial and spectral transformations also appear in the context of hyperspectral super-resolution \cite{KaFuSiMa18}.

In order to address the challenges in data fusion applications, in addition to more general couplings, the formulation in (\ref{eq:cmtf}) needs to be further extended to incorporate various constraints and loss functions. In many applications, it is necessary to impose certain constraints or regularizations on the latent factors to obtain physically meaningful and identifiable patterns. Popular constraints include non-negativity \cite{EyHuDe17,ErAcCe13,YoKiKaCh10}, sparsity \cite{AcGuRa12,DeWiBeAn11} and smoothness \cite{TiKi02}. Furthermore, while Frobenius norm-based loss functions have so far been common in coupled factorizations, this loss function implicitly assumes that entries in the data tensor $\tT_i$ are normally distributed around their mean parameterized by the CPD model. However, for nonnegative, discrete or binary data, this assumption is usually not valid and other loss functions yield more accurate results \cite{LiSuCaKo09, HoKoDu19,ErAcCe13,SiGo08a}.

Existing algorithmic approaches for coupled matrix and tensor factorizations are usually only able to handle a few special constraints, exact couplings and Frobenius norm-based loss functions. The CMTF Toolbox \cite{AcKoDu11b} which uses all-at-once gradient-based optimization with quasi-Newton methods can only handle box-constraints. Some other constraints and regularizations are possible but require the algorithm to be redesigned.  The same holds in principle for Gauss-Newton type methods as implemented in Tensorlab~\cite{SoBaLa15, Vervliet2016Tensorlab}, where nevertheless a variety of constraints are available via a transformation of variables.
While linear couplings are possible also in all gradient-based all-at-once optimization methods, they have systematically only been implemented in Tensorlab~\cite{SoBaLa15, Vervliet2016Tensorlab}.
Gradient-based all-at-once methods can be extended to other differentiable loss functions as proposed in \cite{HoKoDu19} for a single CP decomposition. However, such extensions to coupled factorizations are not yet available.
Another framework for handling general loss functions, derived from the exponential family of distributions, in a coupled setting has been proposed \cite{YiCeSi11}. There, factor matrices are updated alternatingly using a Gauss-Newton method, which results in efficient multiplicative and additive update rules in special cases for non-negative and real data, respectively. This algorithm, however, is limited in terms of constraints and efficient only for special instances. More recently, Huang et al.~\cite{Huang2016flexible} proposed a flexible and efficient framework for constrained matrix and tensor factorizations that seamlessly incorporates a wide variety of constraints, regularizations and loss functions. The framework uses Alternating Optimization (AO), where each subproblem is solved inexactly using the Alternating Direction Method of Multipliers (ADMM) \cite{Boyd2011Distributed}. However, coupled factorizations are not explicitly considered.

In order to address these limitations, in this paper we introduce a general algorithmic framework to solve a large class of constrained linearly coupled matrix-tensor factorization problems, building onto the AO-ADMM framework \cite{Huang2016flexible}. Using numerical experiments on simulated and real data, we demonstrate that the proposed framework is:
\begin{itemize}
\item \textbf{flexible:} A large variety of constraints, regularizations, coupling structures and loss functions can be handled in a plug-and-play fashion.
\item \textbf{accurate:} In the case of Frobenius loss, the accuracy achieved by the proposed framework is comparable to state-of-the-art methods. For other loss functions, our proposed framework achieves more accurate results than the extension of the approach introduced in \cite{Huang2016flexible} to coupled factorizations.
\item \textbf{efficient:} We provide efficient closed form updates for a number of commonly used linear coupling relations in the case of Frobenius loss. We show in extensive numerical experiments on simulated data that our approach achieves competitive performance compared to existing methods for problems with exact coupling, and superior performance on problems with linear couplings.
\end{itemize}

In our preliminary results, we have previously demonstrated the promise of AO-ADMM approach for Frobenius norm loss and exact linear couplings \cite{ScCoAc20}. Here, we provide a more detailed derivation and discussion of the algorithm, and the extension to other loss functions. We also include extensive experiments on synthetic data with different levels of difficulty and using different loss functions, as well as a demonstration on a real dataset.
After establishing some background and notations,  in Section~\ref{sec:problemstatement} we state the general coupled factorization problems we aim to solve.
We briefly explain the basics of ADMM and alternating optimization in Section~\ref{sec:AOADMMprel}, before we derive the AO-ADMM framework for regularized linearly coupled matrix and tensor factorizations in Section~\ref{sec:AOADMMforCMTF}. Details of the algorithm are given in Section~\ref{sec:details}. Numerical experiments on simulated and real data are presented in Section~\ref{sec:experiments}. Finally, limitations of the proposed framework and possible extensions are discussed in Section~\ref{sec:extensions}.

\paragraph{Background and Notation}
Here, we define some important tensor notations and concepts, for a full review we refer to \cite{Kolda2009Tensor}.
We denote tensors by boldface uppercase calligraphic letters $\tT$, matrices by boldface uppercase letters $\mM$, vectors by boldface lowercase letters $\vect{v}$ and scalars by lowercase letters $a$. A tensor is a multidimensional array, each dimension is called a \textit{mode}, and  the number of modes is called the \textit{order}. We denote the rank-$R_i$ CP decomposition of a tensor $\tT_i$ of order $D_i$ and size $n_{1_i}\times n_{2_i}\times ... \times n_{D_i}$ as $\KOp{\mC_{i,d}}_{d=1}^{D_i}$, as defined in \eqref{eq:CP1},
where
$\mC_{i,d} ~\in~ \R^{n_{d_i}\times R_i}$ are called the factor matrices. We use the same notation for the matrix case $D_i=2$, where $\KOp{\mC_{i,1}, \mC_{i,2}}=\mC_{i,1}\mC_{i,2}^T$ corresponds to a generic matrix decomposition. Furthermore, $\ast$ denotes the (element-wise) Hadamard product between two equally sized matrices, $\otimes$ denotes the Kronecker product of two matrices $\mA \in \R^{M\times N}$, $\mB \in \R^{K \times J}$,
\begin{equation*}\small{
\mA \otimes \mB = \begin{bmatrix}
\mA(1,1)\mB & \cdots & \mA(1,N)\mB\\
\vdots & \ddots & \vdots \\
\mA(M,1)\mB & \cdots & \mA(M,N)\mB
\end{bmatrix} \in \R^{MK \times NJ}}
\end{equation*}
and $\odot$ denotes the Khatri-Rao product
\begin{equation*}
\mA \odot \mB = \begin{bmatrix}  \mA(:,1) \otimes \mB(:,1)& \cdots & \mA(:,N) \otimes \mB(:,N)\end{bmatrix}
\end{equation*}
between two matrices with the same number of columns $\mA ~\in~ \R^{M\times N}$, $\mB \in \R^{K \times N}$.
Operator $\vecn(\mC)$ refers to a column-wise vectorization of matrix $\mC$, and ${\tT}_{[d]}$ denotes the mode-$d$ unfolding of tensor $\tT$ into a matrix as defined in \cite{Kolda2009Tensor}. The mode-$d$ unfolding of the CPD model tensor $\tX_i=\KOp{\mC_{i,1}, \mC_{i,2},..., \mC_{i,D_i}}$ has the form ${\tX_i}_{[d]} = \mC_{i,d}\mM_{i,d}^T$ \cite{Kolda2009Tensor}, where
\begin{equation}\label{eq:khatrirao}
\mM_{i,d}:= \mC_{i, D_i}\odot \ldots \odot \mC_{i,d+1}\odot \mC_{i,d-1}\odot \ldots \odot \mC_{i,1}.
\end{equation}
Finally, by $\mI_R$ we denote the $R \times R$ unit matrix.

\section{Regularized CMTF with Linear Couplings}\label{sec:problemstatement}
We start by introducing the formalism needed for the general regularized linearly coupled factorization problems we aim to solve.
We consider $N$ tensors $\left\lbrace\tT_i\right\rbrace_{i=1,...,N}$, of not necessarily equal order $D_i\geq 2$ and size $n_{1_i}\times n_{2_i}\times ... \times n_{D_i}$. Thus, matrices are also included. We suppose that each tensor $\tT_i$ follows approximately a CPD with rank $R_i$,
\begin{equation*}
        \tT_i \approx
         \KOp{\mC_{i,d}}_{d=1}^{D_i},
\end{equation*}
where $\mC_{i,d} \in \R^{n_{d_i}\times R_i}$  denotes the factor matrix of mode $d$ in tensor $\tT_i$.
\begin{figure}[t!]\vspace*{-0.5em}
\centering{\includegraphics[width=0.7\columnwidth]{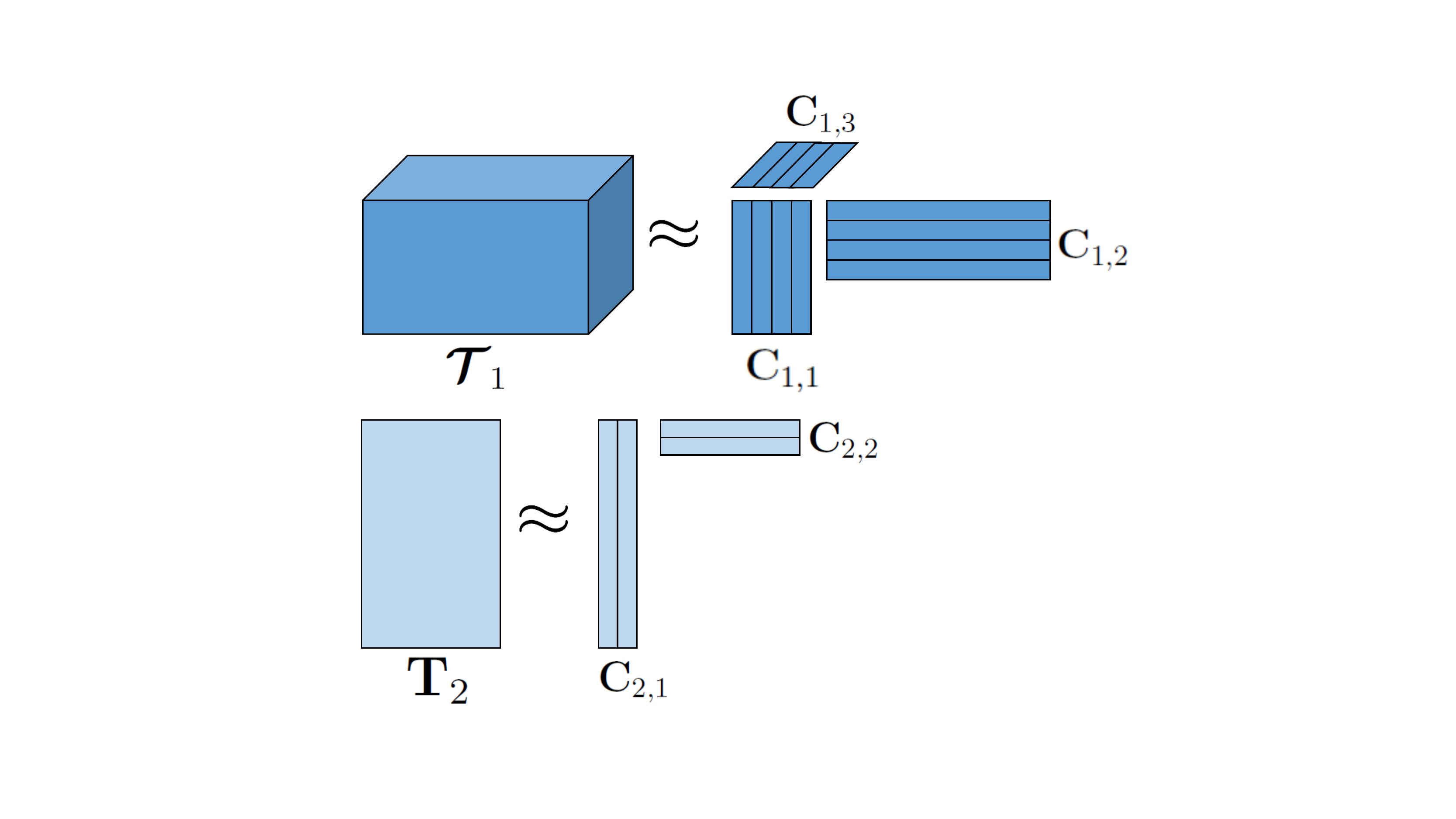}}
\caption{Illustration of data tensors and their decompositions.}
\label{fig:cmtf_notation}\vspace*{-0.5em}
\end{figure}

Moreover, we suppose that some of the factors $\mC_{i,d}$ are
regularized using proper lower semi-continuous convex functions  $g_{i,d}(\mC_{i,d})$.
This covers the important case of constrained factors: suppose factor $\mC_{i,d}$ should belong to a convex set $\mathcal{C}_{i,d}$, then we may set $g_{i,d}=\iota_{\mathcal{C}_{i,d}}$, where $\iota_{\mathcal{C}_{i,d}}$ is the indicator function that is null on $\mathcal{C}_{i,d}$ and infinity elsewhere. Also, $g_{i,d}$ can, for instance, be a sparsity inducing norm such as the $\ell_1$ norm. As we discuss later on, we only require that the proximity operator of $g_{i,d}$ is computable.

Finally, we suppose that some factors are shared across tensors. We consider the case of exact linearly coupled factors, where two or more tensors $\tT_i$ are coupled in any mode $d$ via some underlying matrix $\mDelta_d\in\mathbb{R}^{m_{1,d}\times m_{2,d}}$, as follows:
\begin{equation}\label{eq:gencoupling}
\mH_{i,d} \vecn(\mC_{i,d}) = \mH_{i,d}^{\Delta} \vecn(\mDelta_d), \hspace{0.5cm} i=1,...,N,
\end{equation}
for given transformation matrices $\mH_{i,d}\in\mathbb{R}^{h_{i,d}\times R_in_{d_i}}$ and $\mH_{i,d}^{\Delta}\in~\mathbb{R}^{h_{i,d}\times m_{1,d}m_{2,d}}$. This type of coupling includes many useful cases such as the coupling of modes that have different sampling rates or the partial coupling of tensors that only share a subset of components. Illustrations of such coupling instances are given in Section \ref{sec:cpltypes}. While we focus on exact, \textit{i.e., hard}, couplings in this paper, we propose an extension of our algorithmic framework for approximate, \textit{flexible/soft}, couplings \cite{CabralFarias2016Exploring,Rivet2015Multimodal,ChDaEsKoTh18}, in the Supplementary Material \cite{Suppl}. For ease of notation, for the remainder of the paper we assume
that couplings between different tensors are only in the same mode and there does not exist more than one coupling in each mode.

Using general (convex) loss functions $\mathcal{L}_i(\cdot,\cdot)$, we aim at solving optimization problems of the form
\begin{equation}\label{eq:genoptproblem}\small{
\begin{aligned}
    & \underset{
       \left\lbrace\mC_{i,d},\mDelta_d\right\rbrace_{
      d\leq D_i, i\leq N}}{\argmin} & &
  \sum\limits_{i=1}^{N}\left[ w_i  \mathcal{L}_i \left(\tT_i,
  \KOp{\mC_{i,d}}_{d=1}^{D_i}\right)
  + \sum\limits_{d=1}^{D_i} g_{i,d}\left(\mC_{i,d}\right)\right] \\
   & \ \ \ \ \ \ \ \ \text{s.t. } & &
    \mH_{i,d} \vecn(\mC_{i,d}) = \mH_{i,d}^{\Delta} \vecn(\mDelta_d)
\end{aligned} ,}
\end{equation}
where $w_i$ are weighting parameters.

\noindent{\bf Some reductions.}
While problem~\eqref{eq:genoptproblem} may seem intricate, the following particular settings help reducing its complexity.
\begin{itemize}
  \item \textbf{Loss functions}: Usually, loss functions are chosen to be the squared Frobenius norm. However, in some situations other loss functions may be preferable, see Section~\ref{sec:diffloss}.
  \item \textbf{Regularization-free}:
    Note that setting  $g_{i,d}$ to zero gives the regularization-free, \textit{i.e.,} unconstrained, case.
  \item \textbf{Coupling-free}:
   Similarly, setting $\mH_{i,d}, \mH_{j,d}$ and $\mH_{i,d}^{\Delta}, \mH_{j,d}^{\Delta}$ to contain only zeros, means there is no coupling between tensors $i$ and $j$ in mode $d$. Note that for any two coupled matrices $\mC_{i,d}$ and $\mC_{j,d}$ the constraints, if any, should be compatible with each other and with the transformations.
  \item \textbf{Coupled Matrix and Tensor Factorization (CMTF)}: By setting $g_{i,d}=0$ for all $(i,d)$ and $\mH_{i,1} = \mH_{i,1}^{\Delta} = \matr{I}_{n_1\cdot R}$ for $i=1,2$, the coupled factorization in~\eqref{eq:cmtf}, where shared factors in mode $1$ are exactly identical, is obtained \cite{AcKoDu11b}.
\end{itemize}

Problem \eqref{eq:genoptproblem} is non-convex, and typically difficult to solve for
all $i$ and $d$ simultaneously, in particular, when various constraints or regularizations are
imposed on the factors. However, for many choices of $\mathcal{L}$ and $g$, the cost function is convex w.r.t. $\left\lbrace\mC_{i,d} \right\rbrace_{i=1}^N$ with fixed $d$. Therefore, we propose an Alternating Optimization (AO) algorithm, where each subproblem is solved inexactly using ADMM.

\section{AO-ADMM preliminaries}\label{sec:AOADMMprel}
\subsection{ADMM}
Alternating Direction Method of Multipliers (ADMM) is a primal-dual algorithm
that aims at solving convex constrained optimization problems of the form
\begin{equation}\label{eq:optpb2}
        \underset{\mA \vx +\mB \vz = \vc}{\argmin_{\vx,\vz}} f(\vx) + g(\vz),
\end{equation}
where $f$ and $g$ are (extended-real-valued) convex functions and $f$ is smooth~\cite{Boyd2011Distributed}.
ADMM makes use of Dual Ascent on the augmented Lagrangian to find a solution to~\eqref{eq:optpb2}. The augmented Lagrangian is a cost function $L$ obtained by transforming the linear constraints $\mA \vx+\mB \vz=\vc$ into a penalty
term involving a dual variable $\vmu$, here in scaled form~\cite{Boyd2011Distributed}:
\begin{equation*}
        L(\vx,\vz,\vmu) = f(\vx) + g(\vz) + \frac{\rho}{2}\norm{\mA \vx + \mB \vz - \vc + \vmu}_2^2
\end{equation*}
ADMM is based on the duality theory for convex optimization: the augmented Lagrangian should be minimized
for constant $\vmu$,
but the dual function $G(\vmu)=\min_{\vx,\vz}L(\vx,\vz,\vmu)$ should be maximized w.r.t.
$\vmu$. Therefore, ADMM simply alternates between the minimization of $L$ along
variables $\vx,\vz$ and a gradient ascent step to maximize $G(\vmu)$
as described in Algorithm~\ref{alg:admmplain}.
\begin{algorithm}[!t]
        \begin{algorithmic}
            \WHILE{convergence criterion is not met}
            \STATE{$\small{\vx^{(k+1)}=\underset{\vx}{\argmin} f(\vx) + \frac{\rho}{2} \norm{\smash{ \mA\vx +
                    \mB\vz^{(k)} - \vc + \vmu^{(k)}}}_2^2}$}
            \STATE{$\small{\vz^{(k+1)}=\underset{\vz}{\argmin} g(\vz) +~\frac{\rho}{2} \norm{\smash{ \mA\vx^{(k+1)} +
                    \mB\vz - \vc + \vmu^{(k)}} }_2^2}$}
                    \STATE{$\small{\vmu^{(k+1)} = \vmu^{(k)} + \mA\vx^{(k+1)} + \mB\vz^{(k+1)} -\vc }$}
            \STATE{$\small{k = k+1}$}
            \ENDWHILE{}
        \end{algorithmic}
        \caption{Skeleton of scaled-form ADMM~\cite{Boyd2011Distributed}}\label{alg:admmplain}
\end{algorithm}

Following \cite{Boyd2011Distributed}, based on primal and dual feasibility conditions, a reasonable termination criterion is that primal ($\vect{r}^{(k)}$) and dual residuals ($\vect{s}^{(k)}$) must be small, \textit{i.e.,}
\begin{equation}\label{eq:ADMMstopcrit}\begin{aligned}
\norm{\smash{\vect{r}^{(k)}}}_2 &= \norm{\smash{\mA\vx^{(k)}+\mB\vz^{(k)}-\vc}}_2\leq \epsilon^{\text{pri}},\\
\norm{\smash{\vect{s}^{(k)}}}_2 &= \norm{\smash{\rho \mA^T \mB \left(\vz^{(k+1)}-\vz^{(k)} \right)}}_2\leq \epsilon^{\text{dual}},
\end{aligned}
\end{equation}
where $\epsilon^{\text{pri}}>0$ and   $\epsilon^{\text{dual}}>0$ are feasibility tolerances, which can be set as follows:
\begin{equation*}\label{eq:ADMMstoptol}\small{\begin{aligned}
\epsilon^{\text{pri}}&=\sqrt{\text{length}(\vc)}\epsilon^{\text{abs}}+\epsilon^{\text{rel}}\max \left\lbrace \norm{\smash{\mA\vx^{(k)}}}_2,\norm{\smash{\mB\vz^{(k)}}}_2,\norm{\smash{\vc}}_2 \right\rbrace,\\
\epsilon^{\text{dual}}&=\sqrt{\text{length}(\vx)}\epsilon^{\text{abs}}+\epsilon^{\text{rel}}\rho \norm{\smash{\mA^T \vmu^{(k)}}}_2,
\end{aligned}}
\end{equation*}
with  $\epsilon^{\text{abs}}>0$ and $\epsilon^{\text{rel}}>0$ denoting the absolute and relative tolerance, respectively.

ADMM has become popular in recent years because of mainly two features.
First, given mild hypotheses on the convex functions $f$ and $g$~\cite{Boyd2011Distributed}, it is guaranteed to
converge to the optimal solution of~\eqref{eq:optpb2}. Second, in many
particular instances, ADMM can be implemented using parallel computing, which
leverages modern computer architectures.
Moreover, ADMM is built upon the theoretical development of proximal
operators.  For any $\lambda>0$, a proximal operator of a function $g$ is the following
function~\cite{Moreau1965Proximite,Be17}
\begin{equation}\label{eq:proxop}
        \prox_{\lambda g}(\vx) = \underset{\vect{u}}{\argmin} g(\vect{u}) + \frac{1}{2\lambda} \norm{ \vx - \vect{u}
        }_2^2,
\end{equation}
which is single-valued for convex $g$ and well-defined for proper lower semi-continuous $g$.
For many functions $g$ such as indicator functions of convex sets, a closed-form expression
for the proximal operator is available, see lists here ~\cite{proxop,Be17}. One may notice
that ADMM updates for $\vx$ and $\vz$ are directly proximal operators if
respectively $\mA$ or $\mB$ are orthogonal matrices, although since $f$ is differentiable the $\vx^{(k+1)}$ update can be carried out by smooth convex descent algorithms.
\vspace*{-0.5em}
\subsection{AO-ADMM: connecting ADMM with block-coordinate descent.}\label{sec:AO}
Now, suppose the following optimization problem is given:
\begin{equation}
        \label{eq:optpb3}
        \argmin_{\vx,\vy} f(\vx,\vy) + g_{x}(\vx) + g_y(\vy)
\end{equation}
where $f$ is non-convex, but $f_y + g_y: \vy\mapsto f(\vx,\vy) + g(\vy)$ is convex as well as $f_x + g_x$.
A simple yet powerful idea is to partially solve~\eqref{eq:optpb3} w.r.t. $\vx$
for fixed $\vy$, then for $\vy$ with fixed $\vx$, and to iterate this process until
convergence. Namely, the cost function is (approximatively) minimized over blocks $\vx$ and $\vy$
alternatively, an optimization method commonly known as block-coordinate
descent. To solve each convex constrained subproblem, it is
possible to use ADMM described above in a straightforward manner. The resulting
optimization algorithm, that solves alternatively partial problems using ADMM,
has been named as AO-ADMM ~\cite{Huang2016flexible}. The convergence of
AO-ADMM, to the best of our knowledge, is not proven, see also~\cite{Huang2016flexible} for an overview of convergence results.
It is only known, that if the minimum of each subproblem is uniquely attained and the cost is non-increasing on the update path, every limit point of the algorithm is guaranteed to be a stationary point \cite{Bertsekas1999Nonlinear}{ (3rd ed., proposition 3.7.1 pp324)}. To ensure a unique solution of the subproblems, a majorized version can be solved which is strongly convex, the so called block succesive upper-bound minimization (BSUM) \cite{RaHoLu12}. This can, for example, be done by adding a proximal regularization term, with $\alpha_x^{(k)}>0$, such that the update for $\vx$ becomes \cite{Huang2016flexible}
\begin{equation*} \label{eq:bsum}
\vx^{(k+1)}=\argmin_{\vx}f(\vx,\vy^{(k)})+g_x(x) + \frac{\alpha_x^{(k)}}{2}\norm{\vx-\vx^{(k)}}_2^2.
\end{equation*}

\subsection{Related work: ADMM for constrained tensor factorizations}
Besides the already mentioned AO-ADMM framework \cite{Huang2016flexible}, many examples of ADMM-based algorithms for constrained matrix or tensor factorizations can be found in the literature. For instance, the AO-ADMM framework \cite{Huang2016flexible} has been extended to deal with robust tensor factorization, where some slabs are grossly corrupted\cite{FuHuMaSiBr15}. A parallelization strategy and high performance implementation of the AO-ADMM framework \cite{Huang2016flexible} has been presented \cite{SmBeKa17}. Furthermore, ADMM has been used for non-negative matrix factorizations with the beta-divergence \cite{FeBeDu09,SuFe14}, and for fitting CP models with various constraints \cite{LiSi15, WaChCh15}. Recently, Afshar et al. have also used ADMM to fit a constrained PARAFAC2 model \cite{AfPePaSeHoSe18}. Direct ADMM has also been applied to coupled matrix factorizations with $\ell_1$ norm loss in the Robust JIVE algorithm \cite{SaPaLe17}.

\section{AO-ADMM for regularized CMTF with linear couplings}
\label{sec:AOADMMforCMTF}
Using the algorithms described in the previous section, we can now derive a flexible algorithmic framework to solve the general optimization problem for regularized linearly coupled matrix and tensor factorizations~\eqref{eq:genoptproblem}, which is repeated here for convenience:
\begin{equation*}
\small{
\begin{aligned}
    & \underset{
       \left\lbrace\mC_{i,d},\mDelta_d\right\rbrace_{
      d\leq D_i, i\leq N}}{\argmin} & &
  \sum\limits_{i=1}^{N}\left[ w_i  \mathcal{L}_i \left(\tT_i,
  \KOp{\mC_{i,d}}_{d=1}^{D_i}\right)
  + \sum\limits_{d=1}^{D_i} g_{i,d}\left(\mC_{i,d}\right)\right] \\
   & \ \ \ \ \ \ \ \ \text{s.t. } & &
    \mH_{i,d} \vecn(\mC_{i,d}) = \mH_{i,d}^{\Delta} \vecn(\mDelta_d)
\end{aligned}}
\end{equation*}
As mentioned before, this problem is non-convex with respect to all arguments.
However, the problem w.r.t. the block of parameters $\left\lbrace\lbrace\mC_{i,d} \right\rbrace_{i=1}^N,\mDelta_d\rbrace$ with fixed mode $d$, \textit{e.g.,} for mode $1$,
\begin{equation*}\small{
\begin{aligned}
    & \underset{
      \left\lbrace\mC_{i,1}\right\rbrace_{i\leq N},\mDelta_1}{\argmin}
    \hspace*{-0em} & &
  \sum\limits_{i=1}^{N}\left[ w_i  \mathcal{L}_i \left(\tT_i,
  \KOp{\mC_{i,d}}_{d=1}^{D_i}\right)
  + g_{i,1}\left(\mC_{i,1}\right)\right] \\
 & \ \ \ \ \ \  \text{s.t. } & &
    \mH_{i,1} \vecn(\mC_{i,1}) = \mH_{i,1}^{\Delta} \vecn(\mDelta_1)
\end{aligned}}\label{eq:suboptproblem}
\end{equation*}
is convex for convex functions $\mathcal{L}_i$ and $g_{i,1}$. For each factor matrix $\mC_{i,1}$ we define a ``split'' matrix variable $\mZ_{i,1}$, similar to $\vz$ in~\eqref{eq:optpb2}, which separates the regularization from the factorization, but introduces the additional equality constraint $\mC_{i,1}=\mZ_{i,1}$. Variable $\mDelta_1$ can also be seen as a split variable since it decouples the coupled factor matrices of different tensors. This leads to a convex optimization problem similar to~\eqref{eq:optpb2},
\begin{equation}\small{
\begin{aligned}
       & \argmin_{ \left\lbrace\mC_{i,1},\mZ_{i,1}\right\rbrace_{i\leq N},\mDelta_1} & &
        \sum_{i=1}^{N}\left[ w_i  \mathcal{L}_i \left(\tT_i,
        \KOp{\mC_{i,d}}_{d=1}^{D_i}\right)
        +  g_{i,1}\left(\mZ_{i,1}\right)\right]\\
       & \ \ \ \ \ \  \text{s.t.} & & \mH_{i,1} \vecn(\mC_{i,1}) = \mH_{i,1}^{\Delta} \vecn(\mDelta_1)\\
       & \ \ \ \ \ \ & & \mC_{i,1}=\mZ_{i,1}
\end{aligned}}\label{eq:suboptproblemADMM}
\end{equation}
which can be solved with ADMM.
In contrast to \eqref{eq:optpb2} and also \cite{Huang2016flexible}, we use ADMM with three blocks instead of two: $\{\mC_{i,1}\}_{i\leq N}$, $\{\mZ_{i,1}\}_{i\leq N}$ and $\mDelta_1$. Therefore, we introduce two sets of dual variables, $\{{\vmu_{i,1}}_{(z)}\}_{i\leq N}$ (matrix-valued) for the regularization constraint and $\{{\vmu_{i,1}}_{(\delta)}\}_{i\leq N}$ (vector-valued) for the coupling constraint. The augmented Lagrangian can then be written explicitly as follows, with $\vdelta$ denoting $\vecn(\mDelta)$:
\begin{equation*}\label{eq:costfunadmm1}\small{
\begin{aligned}
                 L(\mC_{i,1},\mZ_{i,1},\vdelta_1,{\vmu_{i,1}}_{(z)}&,{\vmu_{i,1}}_{(\delta)})= \sum\limits_{i=1}^{N}\bigg[ w_i \mathcal{L}_i \left({\tT_i},\KOp{\mC_{i,d}}_{d=1}^{D_i}
                 \right)
                + g_{i,1}(\mZ_{i,1})
                + \frac{\rho}{2} \norm{\mC_{i,1} -
                        \mZ_{i,1} + {\vmu_{i,1}}_{(z)}}_F^2\\ & +\frac{\rho}{2} \norm{\mH_{i,1}\vecn(\mC_{i,1}) -
                        \mH_{i,1}^{\Delta}\vdelta_1 + {\vmu_{i,1}}_{(\vdelta)}}_2^2
\bigg]
\end{aligned}}
\end{equation*}
Analogously to Algorithm~\ref{alg:admmplain}, alternating minimization of this augmented Lagrangian with respect to the three blocks $\{\mC_{i,1}\}_{i\leq N}$, $\{\mZ_{i,1}\}_{i\leq N}$ and $\mDelta_1$ and a gradient ascent step for the dual variables $\{{\vmu_{i,1}}_{(z)}\}_{i\leq N}$ and $\{{\vmu_{i,1}}_{(\delta)}\}_{i\leq N}$, results in the ADMM Algorithm~\ref{alg:admm1}.
\begin{algorithm}[!t]
        \begin{algorithmic}[1]
            \WHILE{convergence criterion is not met}

    		\FOR{$\small{i=1,...,N}$}
	            \STATE{$\small{\begin{aligned} \mC_{i,1}^{(k+1)}=&\underset{\mX}{\argmin}  w_i
                      \mathcal{L}_i \left(
	                    \tT_i ,\KOp{\mX, \mC_{i,2},..., \mC_{i,D_i}}  \right)
	                    + \frac{\rho}{2} \norm{ \mX -\mZ_{i,1}^{(k)} + {\vmu_{i,1}}_{(z)}^{(k)} }_F^2
	                     + \frac{\rho}{2}\norm{\mH_{i,1}\vecn(\mX) - \mH_{i,1}^{\Delta}\vdelta_1^{(k)} + {\vmu_{i,1}}_{(\delta)}^{(k)}} ^2_2  \end{aligned}}$}
			\ENDFOR{}

            \STATE{$\small{\vdelta_1^{(k+1)} = \underset{\vz}{\argmin}\sum\limits_{i=1}^{N}\norm{
            \mH_{i,1}\vecn(\mC_{i,1}^{(k+1)}) - \mH_{i,1}^{\Delta}\vz + {\vmu_{i,1}}^{(k)}_{(\delta)}}_2^2 }$}

			\FOR{$\small{i=1,...,N}$}
				\STATE{$\small{\begin{aligned} \mZ_{i,1}^{(k+1)}&=\underset{\mZ}{\argmin} g_{i,1}(\mZ)+\frac{\rho}{2}\norm{\mC_{i,1}^{(k+1)}-\mZ+{\vmu_{i,1}}_{(z)}^{(k)}}_F^2 \\ &=
		            \prox_{\frac{1}{\rho}g_{i,1}}\left(\mC_{i,1}^{(k+1)} +
		            {\vmu_{i,1}}_{(z)}^{(k)}  \right)\end{aligned}} $}
	            \STATE{$\small{ {\vmu_{i,1}}_{(z)}^{(k+1)} = {\vmu_{i,1}}_{(z)}^{(k)} +
	            \mC_{i,1}^{(k+1)} - \mZ_{i,1}^{(k+1)} }$}

	            \STATE{$\small{ {\vmu_{i,1}}_{(\delta)}^{(k+1)} =
	                    {\vmu_{i,1}}_{(\delta)}^{(k)} +
	            \mH_{i,1}\vecn(\mC_{i,1}^{(k+1)}) - \mH_{i,1}^{\Delta}\vdelta_1^{(k+1)} }$}
    		\ENDFOR{}
            \STATE{$\small{k = k+1}$}
            \ENDWHILE{}
        \end{algorithmic}
        \caption{ADMM for subproblem w.r.t. mode $1$ of regularized linearly coupled CPD}
\label{alg:admm1}
\end{algorithm}
Note that both \emph{for} loops can be computed using parallel programming. For unconstrained factor matrices $\mC_{i,1}$, all terms involving $\mZ_{i,1}$ become unnecessary and are omitted. Similarly, uncoupled, but constrained, factor matrices $\mC_{i,1}$ are updated in an independent ADMM loop, where all terms involving $\vdelta_1$ are omitted. Finally, for uncoupled and unconstrained factor matrices $\mC_{i,1}$, ADMM iterations are not necessary, since the exact solution can be obtained by a least-squares update. Note that Algorithm~\ref{alg:admm1} is presented for mode $1$ only, for ease of notation, but can easily be adapted to any other mode. Using alternating optimization (AO) as explained in Section \ref{sec:AO}, Algorithm~\ref{alg:admm1} is then repeatedly looped over all modes $d$ to estimate all the variables $\mC_{i,d}$.
This alternating optimization constitutes the frame of our algorithm and is illustrated in Algorithm~\ref{alg:AO_CMTF}.
\begin{algorithm}[!t]
        \begin{algorithmic}
        	\STATE{initialize $\small{\{\mC_{i,d}\}_{d\leq D_i,i\leq N}}$, $\small{\{\mDelta_d\}_{d\leq \max \left(D_i\right)}}$, $\small{\{\mZ_{i,d}\}_{d\leq D_i,i\leq N}}$, $\small{\{{\vmu_{i,d}}_{(\vdelta)}\}_{d\leq D_i,i\leq N}}$, $\small{\{{\vmu_{i,d}}_{(z)}\}_{d\leq D_i,i\leq N}}$}
            \WHILE{convergence criterion is not met}
    		\FOR{$\small{d=1,...,\max_i \left(D_i\right)}$}
	            \STATE{update $\small{\{\mC_{i,d}\}_{i\leq N}}$, $\small{\mDelta_d}$, $\small{\{\mZ_{i,d}\}_{i\leq N}}$, $\small{\{{\vmu_{i,d}}_{(\vdelta)}\}_{i\leq N}}$, $\small{\{{\vmu_{i,d}}_{(z)}\}_{i\leq N}}$, using few iterations of ADMM Algorithm~\ref{alg:admm1}}
			\ENDFOR{}
            \ENDWHILE{}
            \STATE{return factor matrices $\small{\{\mC_{i,d}\}_{d\leq D_i,i\leq N}}$}
        \end{algorithmic}
        \caption{AO-ADMM algorithm for regularized coupled CPD}\label{alg:AO_CMTF}
\end{algorithm}

The main advantage of ADMM is that it splits the problem~\eqref{eq:suboptproblemADMM} into easier individual problems for $\mC_{i,1}$, $\mZ_{i,1}$ and $\vdelta_1$, respectively, which we will discuss in the following. For the commonly used squared Frobenius norm loss, the update of factor matrices $\mC_{i,1}$ (line $3$ of Algorithm \ref{alg:admm1}) results in the solution of a linear least-squares problem, which can be solved efficiently, provided $\mH_{i,1}$ and $\mH_{i,1}^{\Delta}$ have a specific structure. This is discussed in detail in Section~\ref{sec:cpltypes}. In case of other loss functions than the squared Frobenius norm, we resolve to numerical optimization to update $\mC_{i,1}$, as discussed in Section~\ref{sec:diffloss}. In the case of structured couplings, also the $\vdelta_1$ update (line $5$ of Algorithm \ref{alg:admm1}) can be computed efficiently. We refer to the Supplementary Material \cite{Suppl} for explicit updates. Furthermore, handling the regularization reduces to the computation of the proximal operator $\prox_{\frac{1}{\rho}g_{i,1}}\left(\mC_{i,1}^{(k+1)} +
{\vmu_i}_{(z)}^{(k)}\right)$ to update $\mZ_{i,1}^{(k+1)}$ (line $7$ of Algorithm \ref{alg:admm1}).
For many commonly used regularization functions $g$, the corresponding proximal operator can easily be computed. We list some useful proximal operators in Section~\ref{sec:proxops}.
The implementation of the above algorithm can be found in \cite{AOADMMGit}.

\subsection{Different types of linear coupling}\label{sec:cpltypes}
In the case of squared Frobenius norm loss, the updates for factor matrices $\mC_{i,1}$,
\begin{equation*}\small{
\begin{aligned}
\mC_{i,1}^{(k+1)}=\underset{\mX}{\argmin} & w_i
                      \norm{
	                    {\tT_i}_{[1]} - \mX {\mM_{i,1}^{(k)}}^T}_F^2
	                    + \frac{\rho}{2} \norm{ \mX -\mZ_{i,1}^{(k)} + {\vmu_{i,1}}_{(z)}^{(k)} }_F^2
	                     + \frac{\rho}{2}\norm{\mH_{i,1}\vecn(\mX) - \mH_{i,1}^{\Delta}\vdelta_1^{(k)} + {\vmu_{i,1}}_{(\delta)}^{(k)}} ^2_2,
\end{aligned}}
\end{equation*}
where $\mM_{i,1}$ is defined as in \eqref{eq:khatrirao}, are given by a large linear least-squares problem, which can hardly be computed efficiently.
However, for the following five special forms of linear couplings
$
\mH_{i,1} \vecn(\mC_{i,1})~ = ~\mH_{i,1}^{\Delta} \vecn(\mDelta_1)
$, efficient updates are available.

\subsubsection{Case $1$: Hard coupling (no transformation)}
One way of coupling tensors $i$ and $j$ is simply requiring equality of the factor matrices $\mC_{i,1}=\mC_{j,1}$. This case has been used in formulations of coupled matrix-tensor factorizations and coupled tensor factorizations \cite{Wilderjans2009, AcKoDu11b, YiCeSi11, BeKuPa14, SoDoLa15} as well as their applications in data mining \cite{PaMiSiFa14, JeJe16,ErAcCe13}.
Obviously, coupled factor matrices have to be of the same size and obey the same constraints. The coupling constraint can then simply be written in matrix form as
$\mC_{i,1}=\mDelta_1$.
The update of $\mC_{i,1}$ is obtained by solving the following linear system, where ${\vmu_i}_{(\Delta)}$ denotes the matricized version of ${\vmu_i}_{(\delta)}$,
\begin{equation*}\small{\begin{aligned}
\mC_{i,1}^{(k+1)} &\left[w_i {\mM_{i,1}^{(k)}}^T\mM_{i,1}^{(k)} +\frac{\rho}{2}\left(\mI_{R}+\mI_{R}\right)\right]=
    \left[w_i{\tT_i}_{[1]}\mM_{i,1}^{(k)} +\frac{\rho}{2}\left(\mZ_{i,1}^{(k)}-{\vmu_{i,1}}^{(k)}_{(z)}+\mDelta_1^{(k)} - {\vmu_{i,1}}^{(k)}_{(\Delta)}\right) \right].
\end{aligned}}
\end{equation*}

\subsubsection{Case $2$: Transformations in mode dimension}
Often, measurements obtained from different instruments will correspond to different temporal or spatial sampling grids or aggregation intervals. For instance, assume tensors $i$ and $j$ have different dimensions $n_{1_i}$, $n_{1_j}$ in mode $1$, due to different sampling rates, but sample the same underlying function. Although direct coupling would not make any sense, it may still be possible to approximate the common underlying function via interpolations. Those interpolations can then be compared on a common sampling grid of size $n_{\Delta_1}$ \cite{FaCo15}, where the consensus variable $\mDelta_1 \in \R^{n_{\Delta_1} \times R}$ represents the function on the common grid. There are two possibilities for such couplings. Illustrations of these are shown in Fig.~\ref{fig:coupling2}. We discuss rank and size restrictions on the particular transformation matrices in the Supplementary Material \cite{Suppl}.

\paragraph{Case $2$a}
The factor matrices can be coupled via known transformation matrices $\tilde{\mH}_{i,1} \in \R^{n_{\Delta_1} \times n_{1_i}}$, $n_{\Delta_1} \leq \min_i n_{1_i}$, such that
\begin{equation*}
\tilde{\mH}_{i,1}\mC_{i,1}=\mDelta_1.
\end{equation*}
This results from setting $\mH_{i,1}^{\Delta}=\mI_{n_{\Delta_1}\cdot R}$ and $\mH_{i,1}=\mI_R \otimes \tilde{\mH}_{i,1}$. The update for $\mC_{i,1}$ is then given by the solution of the following Sylvester equation
\begin{equation*}\label{eq:C_case2a}\begin{aligned}
&\frac{\rho}{2}\left(\mI_R+ \tilde{\mH}_{i,1}^T\tilde{\mH}_{i,1} \right)\mC_{i,1}^{(k+1)} +\mC_{i,1}^{(k+1)}w_i{\mM_{i,1}^{(k)}}^T\mM_{i,1}^{(k)}
=w_i{\tT_i}_{[1]}\mM_{i,1}^{(k)}+\frac{\rho}{2}\left[\mZ_{i,1}^{(k)}-{\vmu_{i,1}}^{(k)}_{(z)}+\tilde{\mH}_{i,1}^T \left(\mDelta_1^{(k)} - {\vmu_{i,1}}^{(k)}_{(\Delta)}\right)\right].
\end{aligned}
\end{equation*}
Examples of this type of coupling can be found in \cite{EyHuDe17}, where a model of neurovascular coupling and downsampling is exploited to deal with the different temporal resolutions of EEG and fMRI. In \cite{ChDaEsKoTh18} also the spatial mode of EEG and fMRI is matched using the lead field matrix. In~\cite{Rivet2015Multimodal}, the transformation matrices correspond to first or second order (numerical) derivative matrices.
Note that in \cite{EyHuDe17} the matrix $\tilde{\mH}_{i,1}$ is actually a parameterized model of a hemodynamic response function and the parameters are fitted in the optimization process. We leave learning $\tilde{\mH}_{i,1}$ as future work.

\paragraph{Case $2$b}
Transformations in Case $2$a do not allow the common sampling grid to have more points $n_{\Delta_1}$ than the smallest dimension $n_{1_i}$ of all coupled factor matrices. This restriction can be avoided by transformations of the form
\begin{equation*}
\mC_{i,1}=\tilde{\mH}_{i,1}^{\Delta}\mDelta_1,
\end{equation*}
where $\tilde{\mH}_{i,1}^{\Delta}$ is of size $n_{1_i} \times n_{\Delta_1}$. This is obtained by setting $\mH_{i,1}=\mI_{n_{1_i}\cdot R}$ and $\mH_{i,1}^{\Delta}=\mI_{R} \otimes \tilde{\mH}_{i,1}^{\Delta}$.
The updates of the factor matrices result in solving the following linear systems
\begin{equation*}\small{\begin{aligned}
\mC_{i,1}^{(k+1)} &\left[w_i {\mM_{i,1}^{(k)}}^T\mM_{i,1}^{(k)}+\frac{\rho}{2}\left(\mI_{R}+ \mI_{R} \right)\right] =
 \left[w_i{\tT_i}_{[1]}\mM_{i,1}^{(k)} +\frac{\rho}{2}\left(\mZ_{i,1}^{(k)}-{\vmu_{i,1}}^{(k)}_{(z)}+\tilde{\mH}_{i,1}^{\Delta} \mDelta_1^{(k)} - {\vmu_{i,1}}^{(k)}_{(\Delta)}\right) \right].
\end{aligned}}
\end{equation*}
An example of such a formulation can be found in the context of tensor disaggregation \cite{AlKaSi20}, where the transformation corresponds to an aggregation matrix. It is also used in a tensor fusion model for hyperspectral super-resolution, where the first transformation corresponds to blurring and downsampling to model the spatial degragation in the hyperspectral image and the second transformation corresponds to band-selection and averaging for spectral degragation in the multispectral image \cite{KaFuSiMa18}.
\begin{figure}[t!]
\centering{\includegraphics[width=\columnwidth]{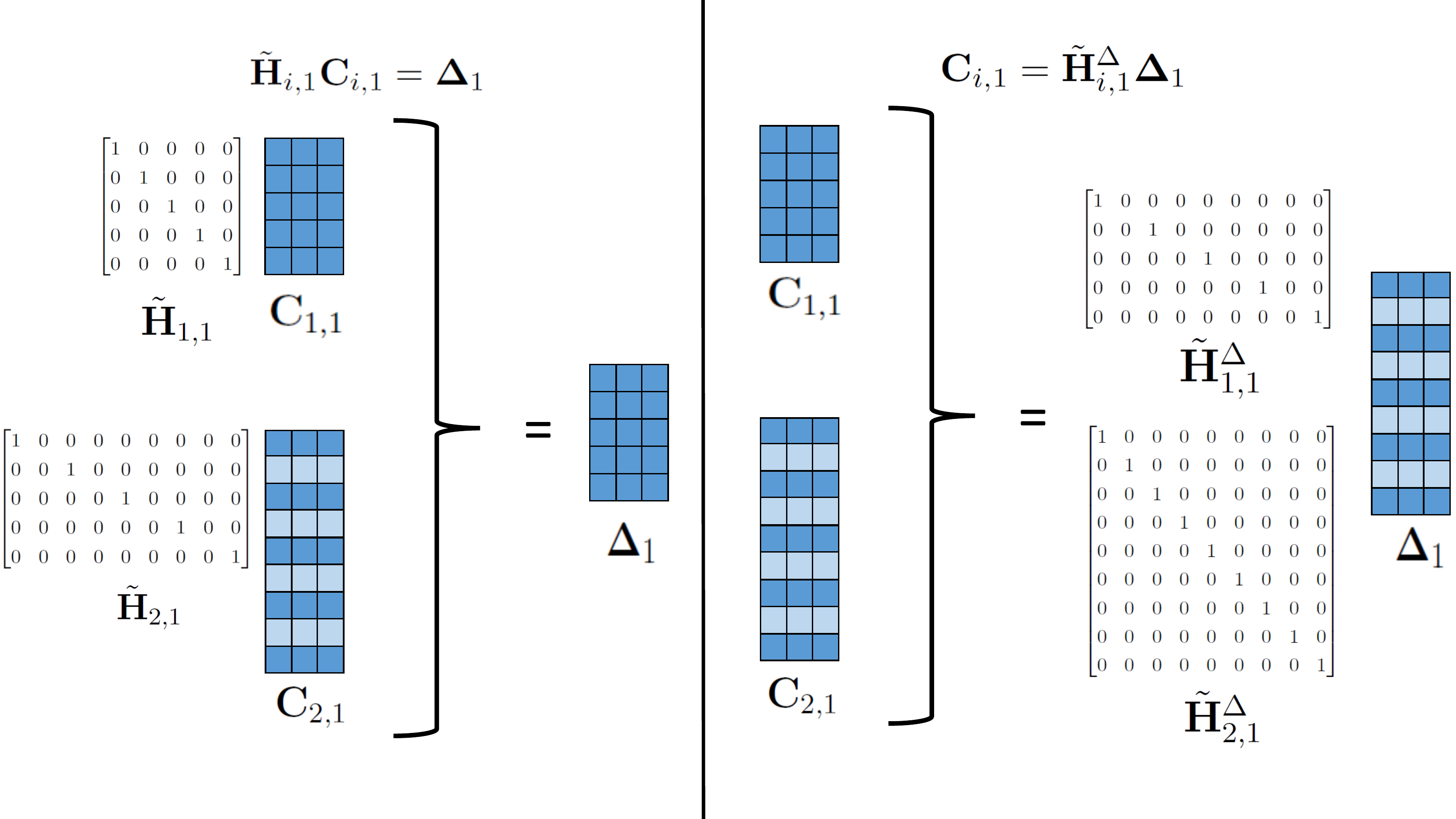}}
\caption{Illustration of possible couplings in Case $2$ where tensors $1$ and $2$ have different dimensions in mode $1$ and the factor matrix $\mC_{1,1}$ contains only every second row of matrix $\mC_{2,1}$. Left: Case $2$a. Right: Case $2$b.}
\label{fig:coupling2}
\end{figure}

\subsubsection{Case $3$: Transformations in component dimension}
\begin{figure}[t!]
\centering{\includegraphics[width=\columnwidth]{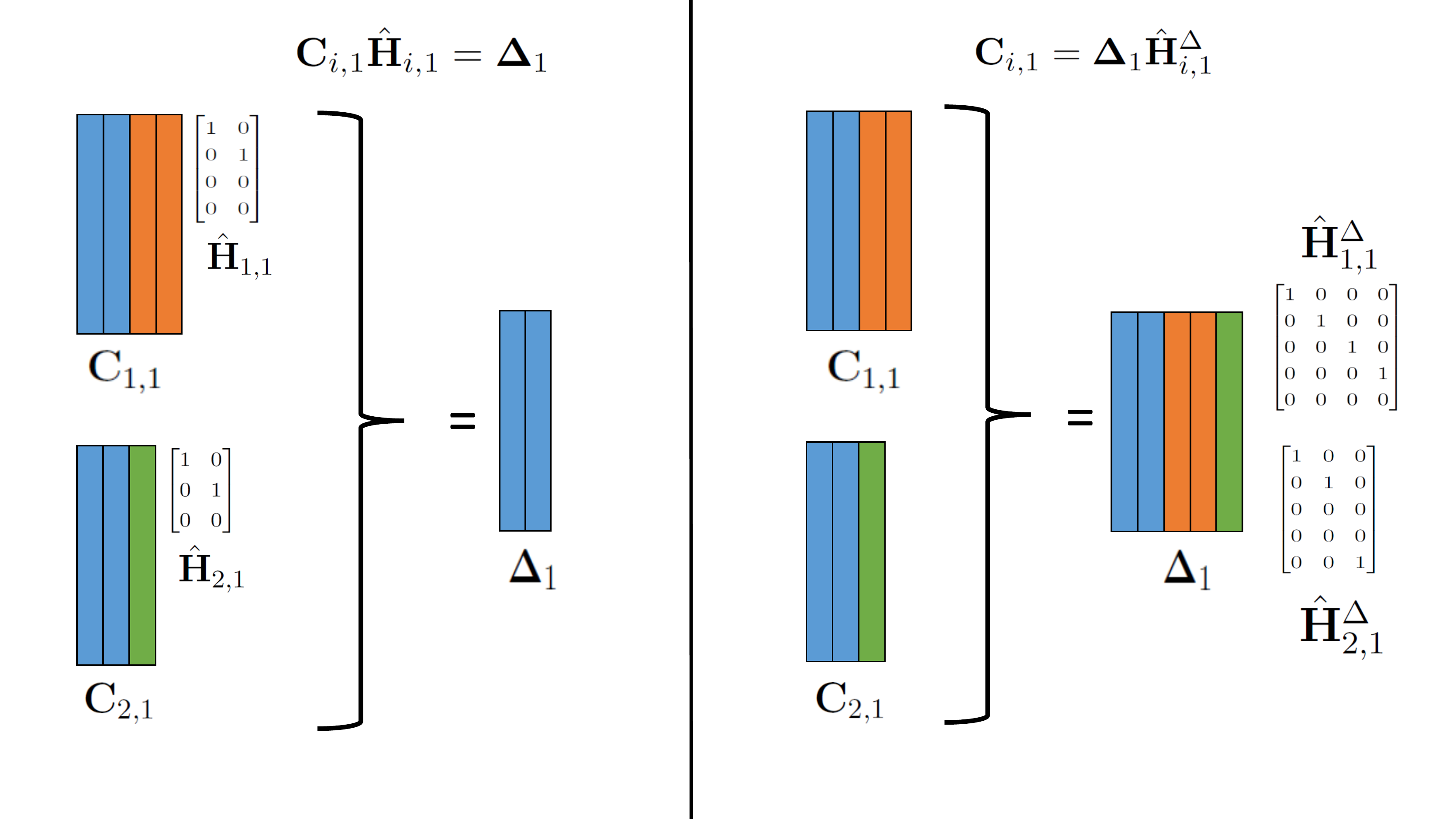}}
\caption{Illustration of possible coupling where two components are shared between tensor $1$ and $2$, tensor $1$ has two additional components, tensor $2$ has one additional component. Left: Case $3$a. Right: Case $3$b.}
\label{fig:coupling3}
\end{figure}

Transformations in the component dimension can be used in cases where coupled tensors do not have the same rank and/or share only a part of their components. We again differentiate between two subtypes.
\paragraph{Case $3$a}
Setting $\mH_{i,1}^{\Delta}=\mI_{n_1\cdot R_\Delta}$ and $\mH_{i,1}=\hat{\mH}_{i,1}^T \otimes \mI_{n_1}$ results in linear couplings of the form
\begin{equation}
\mC_{i,1}\hat{\mH}_{i,1}=\mDelta_1,
\end{equation}
with known transformation matrices $\hat{\mH}_{i,1} \in \R^{R_i \times R_{\Delta}}$. This is equivalent to matching linear combinations of the factor vectors of different components via the consensus variable $\mDelta_1 \in \R^{n_1 \times R_{\Delta}}$. It allows tensors with the same size $n_{1_i}=n_{1_j}=n_1$ in mode $1$, but different number of components $R_i \neq R_j$, to be coupled.
In particular, the special case of rectangular ``identity" matrices $\hat{\mH}_{i,1} \in \R^{R_i \times R_{\Delta}}$, where the upper block consists of the $R_{\Delta} \times R_{\Delta}$ unit matrix and the lower block contains only zeros has a meaningful application. In that case, the consensus variable $\mDelta_1$ contains the $R_{\Delta}$ shared vectors of the possible larger factor matrices $\left\lbrace\mC_{i,1}\right\rbrace_{i\leq N}$. The number of those shared components has to be known beforehand. An illustration of this can be seen in Fig.~\ref{fig:coupling3} on the left side.
The update of $\mC_{i,1}$ is then given by
\begin{equation*}\small{\begin{aligned}
&\mC_{i,1}^{(k+1)} \left[w_i {\mM_{i,1}^{(k)}}^T \mM_{i,1}^{(k)}+\frac{\rho}{2}\left(\mI_{R_i}+ \hat{\mH}_{i,1}\hat{\mH}_{i,1}^T \right)\right] =
\left[w_i{\tT_i}_{[1]}\mM_{i,1}^{(k)} +\frac{\rho}{2}\left(\mZ_{i,1}^{(k)}-{\vmu_{i,1}}^{(k)}_{(z)}+\left(\mDelta_1^{(k)} - {\vmu_{i,1}}^{(k)}_{(\Delta)}\right)\hat{\mH}_{i,1}^T\right) \right].
\end{aligned}}
\end{equation*}
Such coupled tensor factorizations with known number of shared and unshared factors have been previously studied \cite{YoCiYa12} and also used to jointly analyze EEG and fMRI signals \cite{ChDaEsKoTh18}.

\paragraph{Case $3$b}
On the other hand, in couplings of the form
\begin{equation*}
\mC_{i,1}=\mDelta_1 \hat{\mH}_{i,1}^{\Delta},
\end{equation*}
where $\mH_{i,1}=\mI_{n_1\cdot R_i}$ and $\mH_{i,1}^{\Delta}=\hat{\mH}_{i,1}^{{\Delta}^T} \otimes \mI_{n_1}$,
the consensus variable $\mDelta_1 \in \R^{n_{1} \times R_{\Delta}}$ can be thought of as a dictionary from which factor matrices are built via known linear combinations of its elements.
Especially, when some components are only shared between
some but not all coupled tensors, $\mDelta_1$ may contain all columns of all coupled factor matrices $\{\mC_{i,1}\}$, no matter if they belong to shared or unshared components. This information can be encoded in ``identity" matrices $\hat{\mH}_{i,1}^{\Delta} \in \R^{R_{\Delta} \times R_i}$, which contain exactly one $1$ per column and at most one $1$ per row. An illustration of this can be seen in Fig.~\ref{fig:coupling3} on the right side. The update of the factor matrices is given by
\begin{equation*}\small{\begin{aligned}
\mC_{i,1}^{(k+1)}& \left[w_i{\mM_{i,1}^{(k)}}^T\mM_{i,1}^{(k)}+\frac{\rho}{2}\left(\mI_{R_i}+ \mI_{R_i} \right)\right] =
\left[w_i{\tT_i}_{[1]}\mM_{i,1}^{(k)} +\frac{\rho}{2}\left(\mZ_{i,1}^{(k)}-{\vmu_{i,1}}^{(k)}_{(z)}+ \mDelta_1^{(k)}\hat{\mH}_{i,1}^{\Delta} - {\vmu_{i,1}}^{(k)}_{(\Delta)}\right) \right].
\end{aligned}}
\end{equation*}
An example of this type of coupling has been used in audio source separation \cite{SiErCeAc13}, where a binary transformation matrix is used to make sure that only a predefined number of columns of a spectral template are used in one of the decompositions.

Additionally, for the combination of coupling Cases $2$a with $3$b and $2$b with $3$a, efficient updates with small linear systems or Sylvester equations, respectively, can be derived in a similar way.

\subsection{Different loss functions}
\label{sec:diffloss}
Various data mining applications may require cost functions other than the Frobenius norm, which relies on the underlying assumption that the data follows a Gaussian distribution. Social network analysis often relies on count data, which does not follow the Gaussian assumption; therefore, other loss functions such as the Kullback-Leibler (KL) divergence have proven to be more effective for this kind of data. Similarly, if $\tT$ contains music pieces, an Itakuro-Saito (IS) divergence-based cost function has shown to reveal more interpretable results \cite{FeBeDu09}. From a statistical perspective, the cost function is a by-product of the distribution of the entries of $\tT_i$ around their mean parameterized by the CPD model, recently studied in detail in \cite{HoKoDu19}. Therefore, it is possible to imagine a situation where various $\tT_i$ have various data fitting functions.
For a general loss function $\mathcal{L}_i(\cdot,\cdot)$, the  minimization problem in line $3$ of Algorithm~\ref{alg:admm1}
may not have a closed form solution. Therefore, we resolve to numerical optimization. We use the limited-memory BFGS with bound constraints (L-BFGS-B) \cite{ByLuNoZh} in the implementation \cite{BeckerLBFGSB}. Using L-BFGS-B has the advantage that bound constraints, like non-negativity which is necessary for several loss functions, can be ensured in every iteration of the algorithm. The disadvantage of using a gradient based optimization method is that our framework will be limited to loss functions $\mathcal{L}_i(\cdot,\cdot)$ which are differentiable in the second argument. Details about the gradient computation can be found in the Supplementary Material \cite{Suppl}. Nevertheless, many loss functions are differentiable and can be handled by this approach, some important examples are given in Table \ref{tab:lossfunctions} together with references to some of their applications in matrix and tensor factorizations. There, $\ell$ denotes the element-wise loss function, such that \begin{equation}\label{eq:elementwiseLoss}
\mathcal{L}(\tT,\tX)=\sum_{j \in \mathcal{J}} \ell (t_{j},x_{j}),
\end{equation}
where $j$ is a multiindex $j=(j_1,j_2,...,j_D)$ and $\mathcal{J}$ the set of all possible indices. Note that some loss functions introduce non-negativity constraint on the tensor entries in
$\tX=\KOp{ \mC_{i,1}, \mC_{i,2},..., \mC_{i,D_i}}$, which we enforce via non-negativity constraints on all factor matrices. As described before, these bound constraints can be handled by L-BFGS-B, without the need to introduce additional variables $\mZ_{i,1}$.
$\beta$- and $\alpha$-divergences are generalized families of divergences and include  the squared Frobenius norm, KL and IS divergence as special cases. The Huber loss can be thought as a smooth approximation to the $\ell_1$-loss \cite{HoKoDu19}, which is often used for robust fitting when the data is sparingly corrupted by outliers \cite{ChKo10}.
\begin{table}[t!]\small
\caption{Overview of different loss functions}
\begin{center}
\begin{tabular}{|c|c|c|c|}
\hline
loss & distribution & elementwise loss fucntion & references\\
\hline \hline
Kullback-Leibler (KL) div. & Poisson & $\ell(t,x)= x-t\log x +t\log t -t$, $t \in \N, x\geq 0$ & \cite{LiSuCaKo09,ChKo12}\\
\hline
Itakura-Saito (IS) div. & Gamma & $\ell(t,x)=\frac{t}{x} +\log x -\log t -1$, $ t, x> 0$ & \cite{FeBeDu09}\\
\hline
$\beta$-div. &  & $\ell(t,x)=\frac{t^{\beta}}{\beta(\beta-1)}+\frac{x^{\beta}}{\beta}-\frac{tx^{\beta-1}}{\beta-1}$, $\beta \in \R \setminus \left\lbrace 0,1 \right\rbrace$ & \cite{ErAcCe13, SuFe14,CiZdCh07}\\
\hline
$\alpha$-div. &  & $\ell(t,x)=\frac{1}{\alpha(\alpha-1)}\left( t^{\alpha}x^{1-\alpha}-\alpha t+(\alpha-1)x \right)$, $\alpha \in \R \setminus \left\lbrace 0,1 \right\rbrace$& \cite{CiZdCh07,CiPh09}\\
\hline
Huber-loss &  & $
\ell(t,x;d)=\begin{cases}
\left(t-x\right)^2 \ \ \  &\text{if} \abs{t-x}\leq d\\
2d\abs{t-x}-d^2\ \ \ &\text{otherwise}
\end{cases}$ & \cite{SaPaLe17}\\
\hline
\end{tabular}
\label{tab:lossfunctions}
\end{center}
\end{table}
Another approach to solving the problem in line $3$ of Algorithm $2$ (without the coupling) in an AO-ADMM setting is proposed in \cite{Huang2016flexible}. There, another auxiliary tensor variable $\tilde{\tT}_i$ is introduced which is fitted to the data tensor $\tT_i$ via the loss function $\mathcal{L}_i(\cdot,\cdot)$, while $\tilde{\tT}_i=\KOp{\mC_{i,1}, \mC_{i,2},..., \mC_{i,D_i}}$ is treated as another constraint using ADMM. This results in closed form updates for a number of possible loss functions and is thus more efficient than numerical optimization. However, our experiments show that it typically takes a large number of outer iterations before this approach converges. We also observe that it is less accurate than our approach in the case of KL divergence, see Section \ref{sec:exp5}. We suspect that the ``double approximation" first in Frobenius norm and then in the loss function can have negative effects on accuracy. In the following, we refer to this approach as \textit{split AO-ADMM}.

\subsection{Some useful proximal operators}
\label{sec:proxops}
When a factor matrix $\mC_{i,1}$ is regularized, the update of the corresponding auxiliary variable $\mZ_{i,1}$ in Algorithm \ref{alg:admm1} is given by the proximal operator \eqref{eq:proxop}
\begin{equation*}
\mZ_{i,1}^{(k+1)}=\prox_{\frac{1}{\rho}g_{i,1}}\left(\mC_{i,1}^{(k+1)} + {\vmu_{i,1}}_{(z)}^{(k)}  \right)
\end{equation*}
where the function $g_{i,1}$ can be any proper, lower semi-continuous and convex function \cite{Moreau1965Proximite,ProxOpUserGuide,Be17}. Thus, besides hard constraints given by an indicator function $\iota_{\mathcal{C}}$ of a convex set $\mathcal{C}$, soft constraints are also included in this framework via convex regularization functions.
In many cases, closed form solutions and/or efficient algorithms exist, see \cite{Be17,ProxOpUserGuide} and the references therein, and we resort to the implementations collected in $\textit{The Proximity Operator Repository}$ \cite{proxop}.
In Table~\ref{tab1:proxoperators} we recall only a few constraints and regularizations, which are of practical interest, while many others are also possible \cite{Be17,ProxOpUserGuide}. Note that the regularization can be defined as an element-wise function $g(x)$ or act on columns $\vx$ of factor matrices.
As a result of the scaling ambiguity in the CPD model, normalization of the factors may be needed in coupled factorizations. The usual way to tackle norm constraints on the factors is to consider a $\ell_2$ constraint, \emph{e.g.,} $\norm{\mC_{i,1}(:,j)}_2=1$, where $j$ is any column index \cite{AcPaGu14,Rivet2015Multimodal}.
Here, we can also tackle the normalization of the factors in the same way as other constraints by applying the proximal operator of the $\ell_2$ unit ball
on the columns of the split variables $\mZ_{i,1}$. In some cases, it is possible to combine different constraints with one proximal operator, see \cite{Be17,Magoarou2016Flexible}. In practice it is possible to handle also non-convex regularization functions $g$, as long as \textit{a} solution of the corresponding proximal operator can be computed, since its solution may not be unique.
\begin{table*}[t!]
\caption{Some proximal operators}
\begin{center}
\vspace*{-0.5em}
\begin{tabular}{|c|c|c|}
\hline
structure & function $g(x)$ & proximal operator $\prox_{\frac{1}{\rho}g}(x)$   \\
\hline \hline
non-negativity & $\iota_{\R_+}(x)$ & $\max \left\lbrace 0,x\right\rbrace$  \\
\hline
box-constraints & $\iota_{\text{Box}[\ell,u]}(x)$ & $\min\left\lbrace \max \left\lbrace x,\ell\right\rbrace,u \right\rbrace$ \\
\hline
simplex constraints  & $\iota_{\Delta_N}(\vx)$ & \parbox{10cm}{\centering{$\left[\max\left\lbrace0,x_n-\lambda\right\rbrace \right]_{1\leq n\leq N}$, with $\lambda \in \R$ s.t. $\sum_{n=1}^N \max \left\lbrace0,x_n-\lambda\right\rbrace=1$}} \\
\hline
monotonicity & $\iota_{\left\lbrace x_1 \leq x_2 \leq...\leq x_n \right\rbrace}(\vx)$ & $\left[\max_{1\leq j\leq n} \min_{n\leq k\leq N}\frac{1}{k-j+1}\sum_{\ell=j}^k x_{\ell} \right]_{1\leq n\leq N}$  \\
\hline
hard sparsity constraints & $\iota_{\left\lbrace \vx : \norm{\vx}_1\leq r \right\rbrace}(\vx)$ & \parbox{10cm}{\centering{$ \begin{cases}
\vx, & \norm{\vx}_1\leq r \\
\mathcal{T}_{\lambda^{\star}}(\vx), & \norm{\vx}_1>r,
\end{cases}$\\with soft-thresholding operator $\mathcal{T}_\lambda(x)=\max \left\lbrace 0, \abs{x}-\lambda \right\rbrace \text{sgn}(x)$,\\and $\lambda^{\star}$ is any positive solution of $\norm{\mathcal{T}_{\lambda}(\vx)}_1=r$ }} \\
\hline
normalization & $\iota_{\left\lbrace \vx : \norm{\vx}_2\leq 1 \right\rbrace}(\vx)$ & $1/\left(\max\left\lbrace \norm{\vx}_2,1\right\rbrace \right) \vx$\\
\hline
Lasso regularization & $\gamma\norm{\vx}_1$ & soft-thresholding operator $\mathcal{T}_{\frac{\gamma	}{\rho}}(\vx)$\\
\hline
$\ell_2$-Regularization & $\gamma \norm{\vx}_{2}$ & $\left(1-\frac{\gamma}{\rho \max \left\lbrace \norm{\vx}_2,\gamma/\rho \right\rbrace} \right)\vx$\\
\hline
Smoothness & $\gamma\norm{\mD\vx}_2^2$ & $\left( \frac{2\gamma}{\rho}\mD^T\mD+\mI \right)^{-1}\vx$ \\
\hline
Normalized sparsity & $\iota_{\left\lbrace \vx: \|\vx\|_2=1 \right\rbrace \cap \left\lbrace \vx:\|\vx\|_0\leq k \right\rbrace }$ &  $H_k(\vx)\big/\|H_k(x)\|_2$ where $H_k$ is the hard-thresholding operator~\cite{Magoarou2016Flexible}\\
\hline
\end{tabular}
\label{tab1:proxoperators}
\vspace*{-1.5em}
\end{center}
\end{table*}

\section{Algorithm details}\label{sec:details}
In this section, we present some algorithmic aspects in more detail. For a discussion on efficient implementations, we refer the interested reader to the Supplementary Material \cite{Suppl}.
\subsection{Stopping conditions}
We adapt the stopping criterion \eqref{eq:ADMMstopcrit} of the inner ADMM iterations to Algorithm \ref{alg:admm1}, by splitting it into coupling and constraint related conditions, as follows:
\begin{equation*}\small{\begin{aligned}
&\sum_{i} \left( \|\mC_{i,1}^{(k)}-\mZ_{i,1}^{(k)}\|_F \big/ \|\mC_{i,1}^{(k)}\|_F \right) \leq \epsilon^{\text{p,constr}}\\
&\sum_{i}\left( \|\mH_{i,1} \vecn(\mC_{i,1}^{(k)})-\mH_{i,1}^{\Delta} \vdelta_1^{(k)}\|_2\big/ \|\mH_{i,1}\vecn(\mC_{i,1}^{(k)})\|_2\right)\leq \epsilon^{\text{p,coupl}}\\
&\sum_{i} \left( \|\mZ_{i,1}^{(k+1)}-\mZ_{i,1}^{(k)}\|_F \big/ \|{\vmu_{i,1}}^{(k)}_{(z)}\|_F\right)\leq \epsilon^{\text{d,constr}}\\
&\sum_{i} \left( \|\mH_{i,1}^{\Delta}\left(\vdelta_1^{(k+1)}-\vdelta_1^{(k)}\right)\|_2 \big/ \|{\vmu_{i,1}}^{(k)}_{(\delta)}\|_2\right)\leq \epsilon^{\text{d,coupl}}
\end{aligned}}
\end{equation*}
All of the above conditions should be satisfied to stop the algorithm, except when a predefined maximum number of inner iterations is reached. We usually set this to a reasonably small number between $5$ and $10$, since we do not want to solve the subproblems exactly in the beginning of the algorithm, when the initializations are probably far away from the optimal solution. As the algorithm proceeds, the inner ADMM algorithm typically terminates due to the relative tolerances and the number of iterations goes down. For a study of the maximum number of inner iterations, see also the experiment in Section~\ref{sec:experiment2}.
The whole algorithm is terminated, when each of the following residuals $f^{(k)}_{\star}$,
\begin{equation}\label{eq:residuals}\small{\begin{aligned}
f_{\text{tensors}}^{(k)} &= \sum\limits_{i} w_i  \mathcal{L}_i \left(\tT_i,
        \KOp{\mC_{i,1}, \mC_{i,2},..., \mC_{i,D_i}} \right)\\
f_{\text{couplings}}^{(k)} &= \sum_{i,d}\left(\|\mH_i \vecn(\mC_{i,d}^{(k)})-\mH_i^{\Delta}\vdelta_d^{(k)}\|_2 \big/ \|\mH_i \vecn(\mC_{i,d}^{(k)})\|_2\right)\\
f_{\text{constraints}}^{(k)} &=\sum_{i,d} \left(\|\mC_{i,d}^{(k)}-\mZ_{i,d}^{(k)}\|_F \big/ \|\mC_{i,d}^{(k)}\|_F\right).
\end{aligned}}
\end{equation}
has either reached a small absolute tolerance $\epsilon^{\text{abs,outer}}$, or has not changed more than some small relative tolerance $\epsilon^{\text{rel,outer}}$ compared to the previous iteration,
\begin{equation*}
f^{(k)}_{\star}<\epsilon^{\text{abs,outer}}, \ \ \ \ \
\abs{\smash{ f_{\star}^{(k)}-f_{\star}^{(k-1)}}} \big/ \abs{\smash{f_{\star}^{(k)}}}<\epsilon^{\text{rel,outer}}
\end{equation*}
or a predefined number of maximal outer iterations is reached.
\subsection{Choice of $\rho$}
To the best of our knowledge, an optimal step-size $\rho$ has been derived for ADMM applied to quadratic programming only \cite{GhTeShJo14}. In choosing $\rho$ we follow \cite{Huang2016flexible} and choose a different step-length $\rho_{i,d}^{(k)}$ for the update of each factor matrix $\mC_{i,d}^{(k)}$ as
\begin{equation}
\rho_{i,d}^{(k)}=\|\mM_{i,d}^{(k)}\|_F^2/R_i=\text{trace}({\mM_{i,d}^{(k)}}^T \mM_{i,d}^{(k)} )/R_i,
\end{equation}
which can be efficiently computed, see the Supplementary Material \cite{Suppl}.

\section{Experiments}\label{sec:experiments}
In this section, we assess the performance of the proposed AO-ADMM approach in terms of accuracy and computational efficiency on synthetic datasets, and demonstrate its use on a real dataset. In the case of Frobenius loss, a number of methods is available for comparison. However, their ability to handle specific constraints and coupling structures varies. So whenever applicable, we compare the performance with other commonly used methods, namely Alternating Least Squares (ALS) and all-at-once optimization using quasi-Newton and Gauss-Newton methods. The results show that the AO-ADMM approach achieves comparable performance to state-of-the art methods on problems with hard coupling (Case 1) in a number of settings with varying collinearity, tensor sizes and constraints. Furthermore, our experiments indicate that AO-ADMM has advantages over other methods for problems with linear couplings. Moreover, in the case of KL loss, we show that our AO-ADMM approach is more accurate than the split AO-ADMM approach.
\subsection{Experimental Set-up}
For ALS (referred to as CMTF-ALS), we use our own implementation for CMTF based on cp\_als from the Tensor Toolbox~\cite{TTB_Software}. It can so far only handle hard couplings.
For nonnegativity constraints, we solve the alternating nonnegative least squares problems using Hierarchical ALS (HALS) \cite{Gillis2012Accelerated}. For all-at-once optimization with quasi-Newton methods, we use cmtf\_opt from the CMTF Toolbox~\cite{AcKoDu11b} (referred to as CMTF opt) using Nonlinear Conjugate Gradient (NCG) for unconstrained cases and Limited Memory BFGS with bounds (LBFGS-B) for nonnegativity constraints. The implementation can only handle box-constraints. Linear couplings are theoretically possible, but are not implemented. Finally, for Gauss-Newton, we use the Tensorlab implementation sdf\_nls~\cite{Vervliet2016Tensorlab} (referred to as Tensorlab GN), which can handle a variety of constraints and coupling structures.
Our implementation of the AO-ADMM algorithm can be found in \cite{AOADMMGit}.

We monitor the convergence of different algorithms through the function value $f_{\text{tensors}}$ and the factor match score (FMS). Given the true factor matrices $\mC_{i,d}^{\text{true}}$, the FMS for $\mC_{i,d}$ is computed as
\begin{equation*}\label{eq:fms}
\text{FMS}=\prod_{i=1}^N \frac{1}{R_i}\sum_{r=1}^{R_i} \left( \prod_{d=1}^{D_i} \frac{\left\langle \mC_{i,d}(:,r), \mC_{i,d}^{\text{true}}(:,r)\right\rangle}{\norm{\mC_{i,d}(:,r)}_2 \|\mC_{i,d}^{\text{true}}(:,r)\|_2} \right),
\end{equation*}
after finding the best permutation of factors. We run each algorithm until the relative change in function value is less than $10^{-12}$ or a maximum number of $10.000$ outer iterations has been reached\footnote{Other parameters are set as follows: In cmtf\_opt (NCG/LBFSG-B): \texttt{MaxFuncEvals}/\texttt{maxTotalIts}=$10^5$, \texttt{StopTol}/\texttt{pgtol}=$10^{-32}$, in Tensorlab: \texttt{CGMaxIter}=$15$, \texttt{TolX}=$10^{-32}$, \texttt{TolAbs}=$0$, in HALS: \texttt{maxiter}=$500$.}.
For AO-ADMM, we set the absolute tolerance for the residuals in Eq.~\eqref{eq:residuals} to $10^{-4}$.
For each dataset, five initializations (for the more difficult experiment $4$, ten initializations) are used, and the best run with the lowest final function value is reported. We also report the number of failed runs in Tables~\ref{tab1:failedruns} and \ref{tab1:failedruns_exp5}, where a run is considered a failed run if it reaches the maximum number of iterations or gives an FMS below the threshold $0.99^{\sum_{i=1}^N D_i}$.
When no constraints are imposed and Frobenius loss is used, in the first run, factor matrices $\mC_{i,d}$ are initialized using the first $R$ left singular vectors of the corresponding concatenated (if coupled) and unfolded tensors in mode $d$. Otherwise, factor matrices are initialized at random, drawing from the standard normal or, in the case of nonnegativity, uniform distribution. All dual variables, as well as coupling variables $\mDelta_d$, are initialized using the standard normal distribution. The split variables $\mZ_{i,d}$ are initialized by $\prox_{1,g_{i,d}}\left(\mC_{i,d}\right)$.
The experiments were performed on a standard $16$GB Windows $10$ laptop using MATLAB R2019a, without the use of parallel for-loops in Algorithm~\ref{alg:admm1}. Computing times are measured using the \texttt {tic-toc} function in MATLAB.

\subsection{Simulated Data}
For each experiment, we generate $50$ random datasets. Following the CP model,
tensors $\tX_i$ are constructed from known factor matrices. In the case of Frobenius loss, \textit{i.e.,} experiments $1-4$, tensors $\tT_i$ are then generated by adding noise tensors $\tN_i$ with entries drawn from a standard normal distribution as follows:
\begin{equation*}
\tT_i=\tX_i + 0.2 (\norm{\tX_i}_F \big/ \norm{\tN_i}_F) \tN_i.
\end{equation*}
This corresponds to a signal-to-noise ratio of around $14$dB. We normalize each tensor and set the weights to $w_i~=~1/2$.
In the case of other loss functions, tensors $\tT_i$ are generated by drawing each entry from the underlying distribution, \textit{e.g.,} Poisson distribution for KL-divergence, with the mean given by the corresponding entry in $\tX_i$.

\subsubsection{Experiment 1: Frobenius loss, coupling Case $1$ and unconstrained}
\begin{figure}[!t]
\centering{\includegraphics[width=\columnwidth]{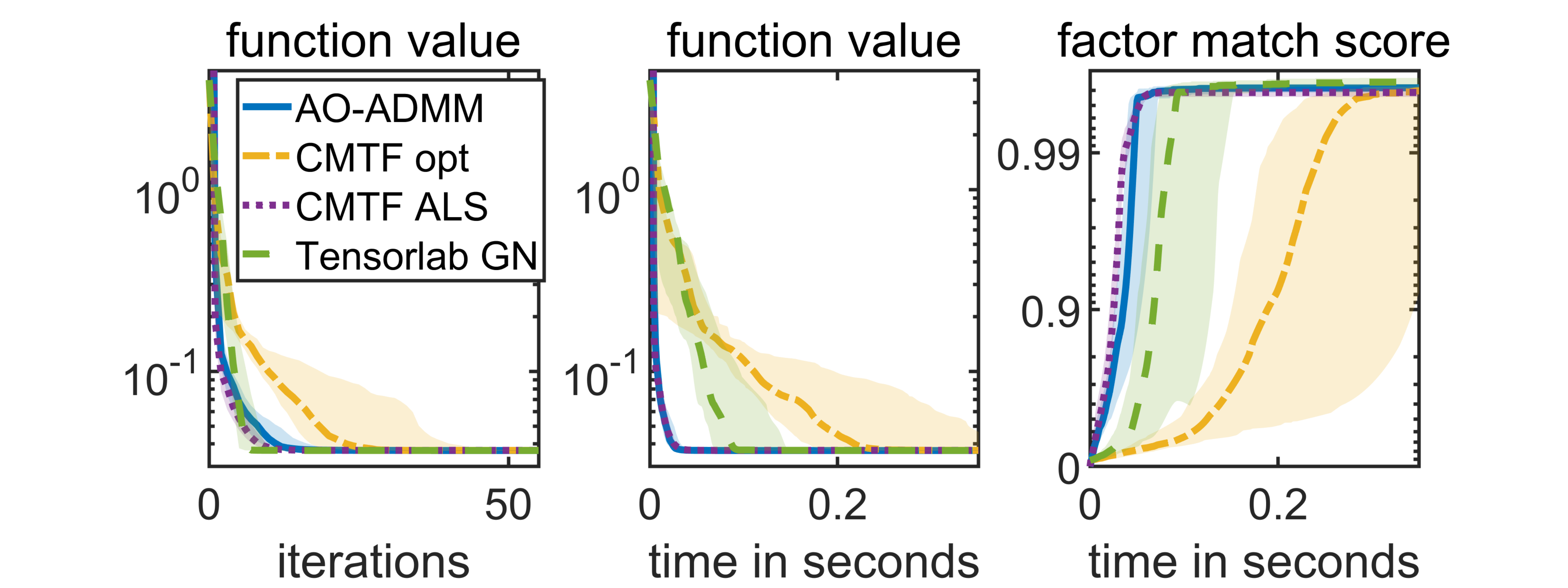}}
\caption{Experiment1a: Median and quartiles of function values and FMS for different algorithms.}
\label{fig:exp1a}
\end{figure}

\paragraph{Mildly Collinear Factors}
\label{sec:Experiment 1a}
In this example, a third-order tensor of size $40 \times 50 \times 60$ is coupled in the first mode with a matrix of size $40 \times 100$ based on hard coupling, \textit{i.e.,} Case $1$. These datasets are jointly analyzed using Frobenius loss and no constraints are imposed. The ground-truth factor matrices with rank $R=3$ are generated with entries drawn from a standard normal distribution and then transformed as in \cite{ToBr06} such that the factors have a congruence of $0.5$.
A summary of the convergence behaviour of different algorithms for $50$ random datasets is shown in Fig. \ref{fig:exp1a}, which depicts the median curves and quantiles of function value and factor match score over iterations and time. Boxplots showing the complete distribution of computing times
can be found in the Supplementary Material \cite{Suppl} for this and the following experiments.
We observe that alternating methods CMTF-ALS and AO-ADMM behave similarly, and are faster than all-at-once optimization methods, as we expect. Also, all methods can accurately recover the true factor matrices by achieving an FMS close to $1$. In the presence of collinearity, singular vectors seem to be a bad choice for initialization for AO-ADMM. Table \ref{tab1:failedruns} shows a high number of failed runs for AO-ADMM which are all due to initializations based on singular vectors. On the other hand, when we use factor matrices with random collinearity, this type of initialization is not an issue (results not shown here). Results for this setting with unbalanced data sizes can be found in the Supplementary Material \cite{Suppl}.

\begin{figure}[!t]
\centering{\includegraphics[width=\columnwidth]{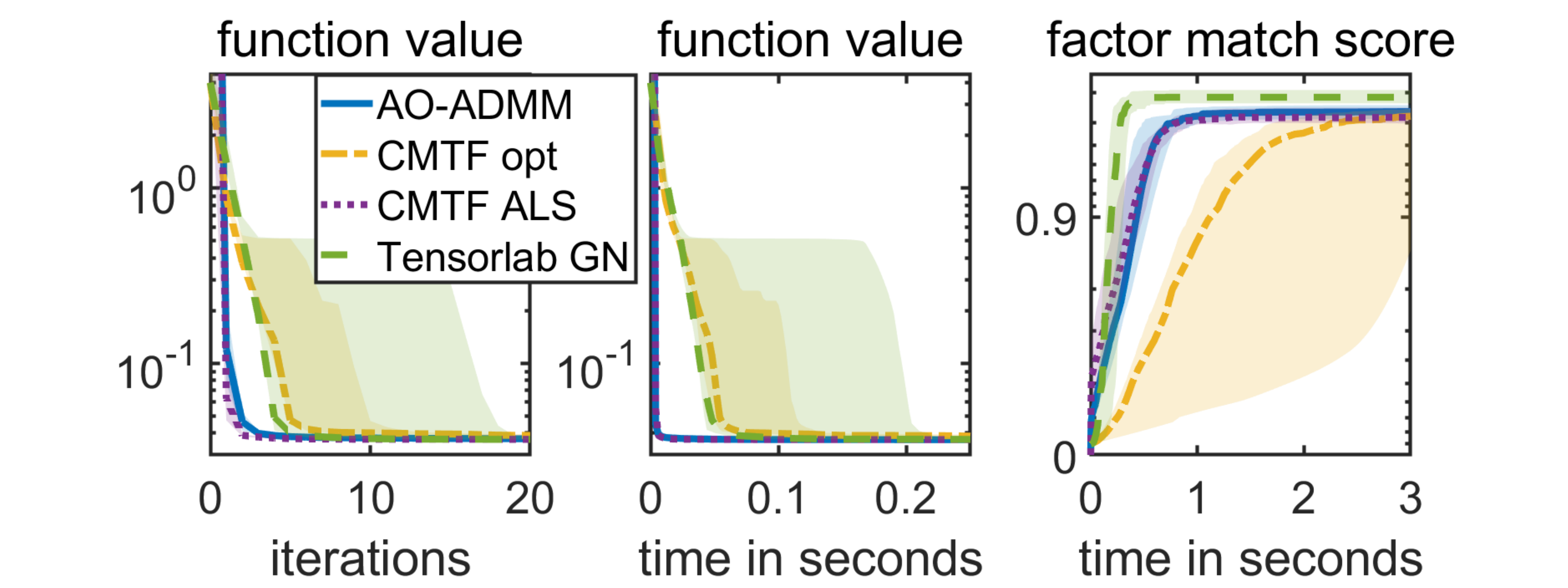}}
\caption{Experiment1b: Median and quartiles of function values and FMS for different algorithms.}
\label{fig:exp1b}
\end{figure}

\paragraph{Highly Collinear Factors}
Here, collinear factors are generated using a congruence of $0.9$. Fig. \ref{fig:exp1b} shows that AO-ADMM again behaves similar to ALS and is also accurate except for the few failing cases shown in Table \ref{tab1:failedruns} for the best 50 runs. In this case, Tensorlab GN is both faster and more accurate, which is expected. In the presence of strong collinearity, ALS type algorithms struggle due to ill-conditioned linear systems. GN type methods, on the other hand, can find an accurate solution fast thanks to the use of approximate second-order information\cite{Vervliet2018Numerical}. Nevertheless, Table \ref{tab1:failedruns} shows that all-at-once optimization-based methods are very sensitive to initializations with a high number of failing runs.

\subsubsection{Experiment 2: Frobenius loss, coupling Case $1$ and nonnegative}\label{sec:experiment2}
\begin{figure}[!t]
\centering{\includegraphics[width=\columnwidth]{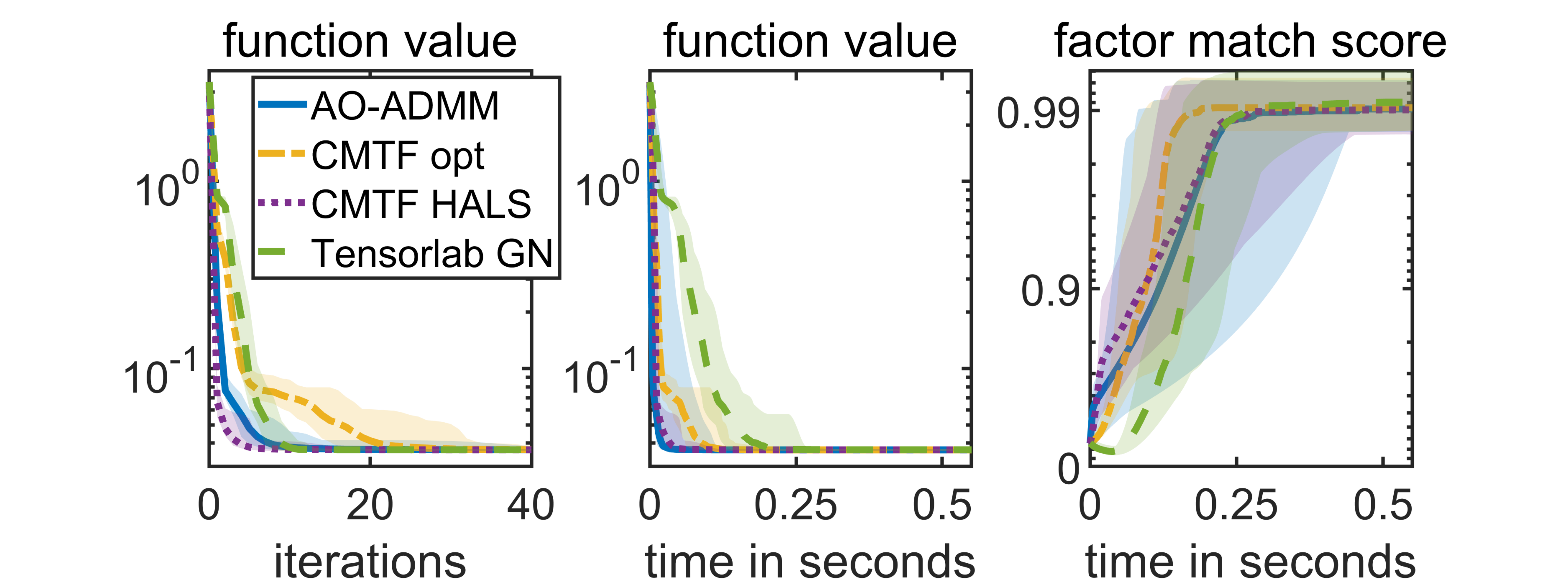}}
\caption{Experiment2: Median and minimum/maximum function values and FMS for different algorithms.}
\label{fig:exp2a}
\end{figure}

\begin{figure}[!t]
\centering{\includegraphics[width=\columnwidth]{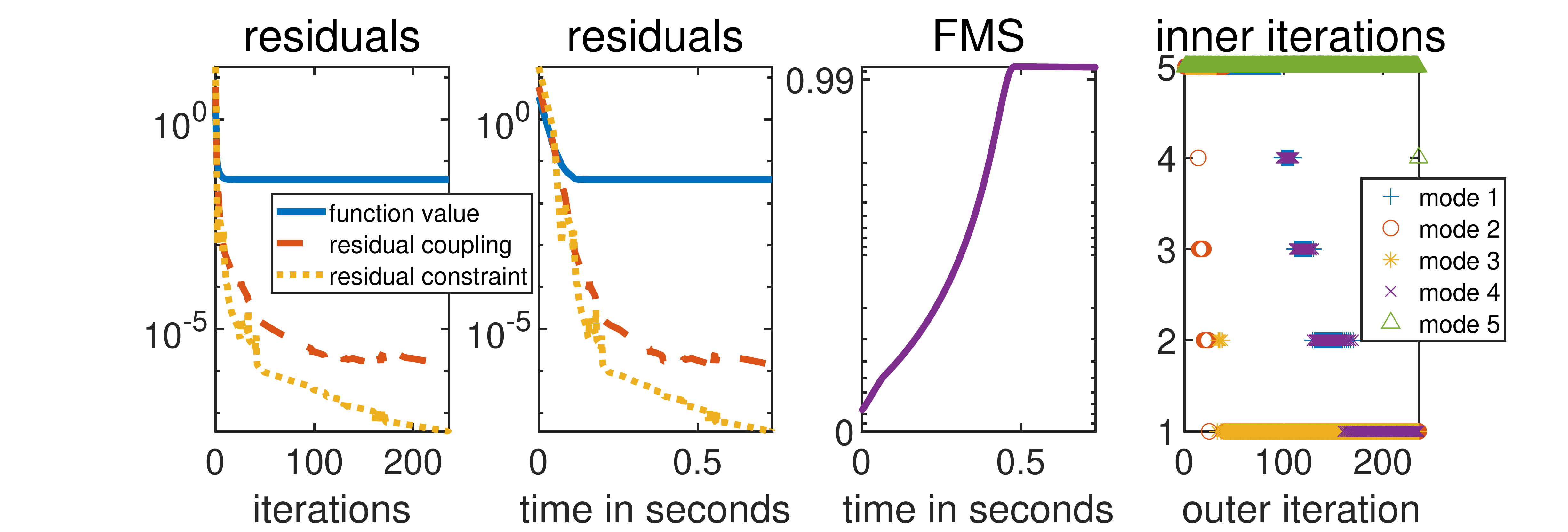}}
\caption{Experiment 2: Detailed convergence of one AO-ADMM run.}
\label{fig:exp2_onlyAOADMM}
\end{figure}

The setting of the second experiment is similar to the first
one. In addition, nonnegativity constraints are imposed on
factor matrices in all modes. The ground-truth factor matrices
are drawn from a uniform distribution and thus the collinearity
of the components is not controlled.
The convergence behaviour of different methods in Fig. \ref{fig:exp2a} demonstrates that
AO-ADMM is computationally efficient, with an average
performance similar to CMTF-HALS.
All methods can accurately recover the true factors used to generate the data.
The convergence of the different residuals in AO-ADMM can be seen in Fig. \ref{fig:exp2_onlyAOADMM} for a single run of AO-ADMM. It can be observed that even though the function does not seem to improve after a number of iterations due to the noise, the residuals for couplings and constraints keep improving, as well as the factor match score. In the right most subfigure, the number of performed inner iterations is plotted against the outer iterations for each mode. All modes start with the maximum of $5$ inner iterations. The iterations go down fast and stay at $1$ for modes $2$ and $3$, which are the uncoupled tensor modes. The coupled modes $1$ and $4$ always have the same number of inner iterations since they are updated together in one ADMM run. Also,  inner iterations of those coupled modes decrease stepwise after some iterations. The only mode that stays at $5$ inner iterations is mode $5$, which is in the presence of noise probably the most difficult one, since it is the uncoupled matrix mode.

Results for unbalanced datasets, as well as a study of different numbers of inner iterations in the AO-ADMM algorithm can be found in the Supplementary Material \cite{Suppl}.

\subsubsection{Experiment 3: Frobenius loss, coupling Case $2$a and unconstrained}
\begin{figure}[!t]
\centering{\includegraphics[width=\columnwidth]{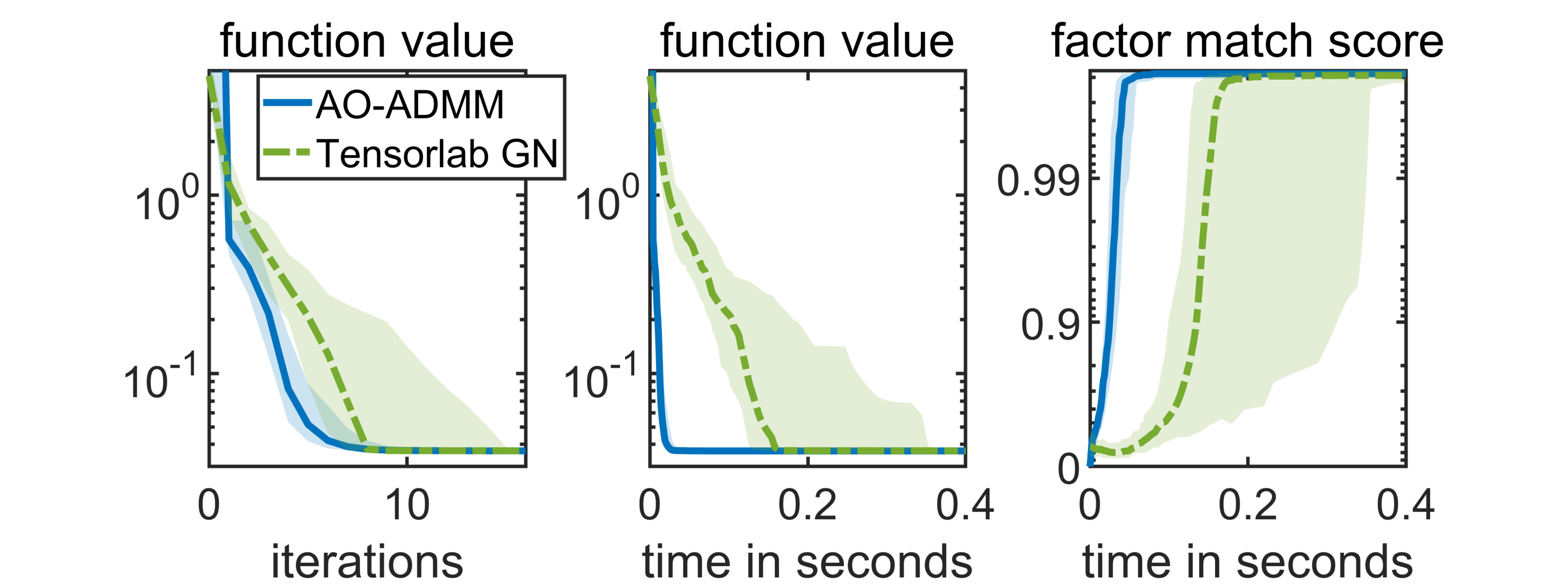}}
\caption{Experiment3: Median and quartiles of function values and FMS for  AO-ADMM and Tensorlab GN.}
\label{fig:exp3}
\end{figure}

In this experiment, we test the linear coupling with transformation on a simple example. Linear couplings of type $2$a allow tensors with different dimensions to be coupled. Here, a tensor $\tT_1$ of size $80\times 50\times 60$ is coupled in the first mode with a matrix $\mT_2$ of size $40 \times 100$ via a transformation $\tilde{\mH}_{1,1} \mC_{1,1}=\Delta_1$, that discards every second entry in $\mC_{1,1}$. The rank of both is again $R=3$ and the true factor matrices have entries drawn from the standard normal distribution.
We compare the AO-ADMM-based approach only with Tensorlab GN since no other implementation is currently able to handle linear couplings. Fig.~\ref{fig:exp3} shows that AO-ADMM finds the true factors faster (on average) than Tensorlab GN. Both are accurate while AO-ADMM seems less sensitive to initialization, see Table~\ref{tab1:failedruns}.

\subsubsection{Experiment 4: Frobenius loss, coupling Case $3$b and unconstrained}
\begin{figure}[!t]
\centering{\includegraphics[width=\columnwidth]{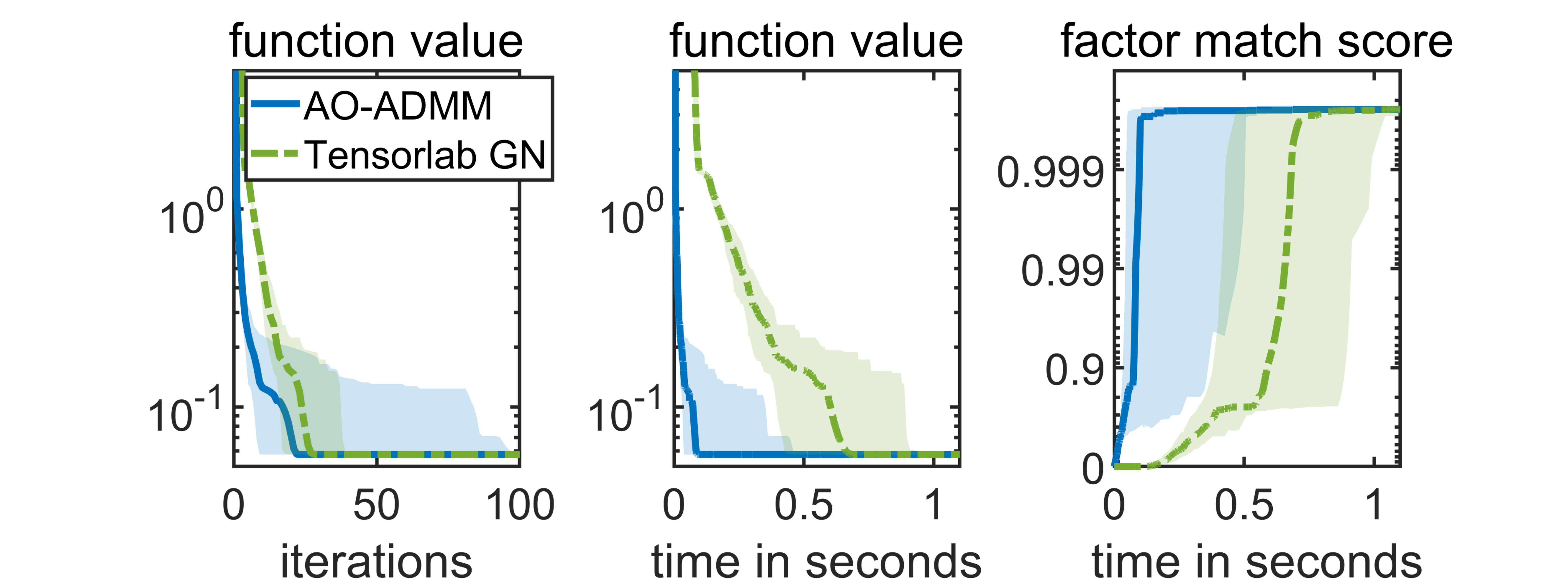}}
\caption{Experiment4: Median and quartiles of function values and FMS for  AO-ADMM and Tensorlab GN.}
\label{fig:exp4}
\end{figure}

Linear couplings of type $3$b allow tensors with different number of components to be coupled. We demonstrate this type of coupling with an example, where three tensors of size $40\times 50\times 60$, $40\times 70\times 60$, $40\times 30\times 50$ and number of components $R=2, 3, 4$, respectively, are coupled in the first mode. Two components are shared by all tensors while the additional component in the second tensor is also present in the third tensor. The third tensor has an additional unshared component. This \textit{ordering} of shared and unshared components makes the problem more difficult and many random initializations will not converge to this order, as shown in Table~\ref{tab1:failedruns}. Therefore, we use $10$ random initializations for each dataset. However, when the best runs out of multiple initializations are considered, both AO-ADMM and Tensorlab GN are accurate while AO-ADMM finds the true factors faster, see Fig.~\ref{fig:exp4}.

\begin{table}[t!]
\caption{Failed runs experiments 1-4}
\begin{center}
\vspace*{-0.5em}
\begin{tabular}{|c|c|c|c|c|c|}
\hline
Exp. &out of & AO-ADMM & CMTF opt & CMTF ALS  & TL GN\\
\hline \hline
 1a & all/ best & 45/0 & 0/0 & 6/0 &12/0\\
\hline
 1b & all/ best & 74/3 & 183/0 & 25/5 &182/1\\
\hline
 2 &all/ best& 0/0 & 0/0 & 0/0 &0/0\\
\hline
 3 & all/ best& 4/0 & - & - & 56/0\\
\hline
 4 & all/ best & 330/1 & - & - & 351/0\\
\hline
\end{tabular}
\label{tab1:failedruns}
\vspace*{-1.5em}
\end{center}
\end{table}

\subsubsection{Experiment 5: KL-divergence and coupling Case $1$}\label{sec:exp5}
\begin{figure}[!t]
\centering{\includegraphics[width=\columnwidth]{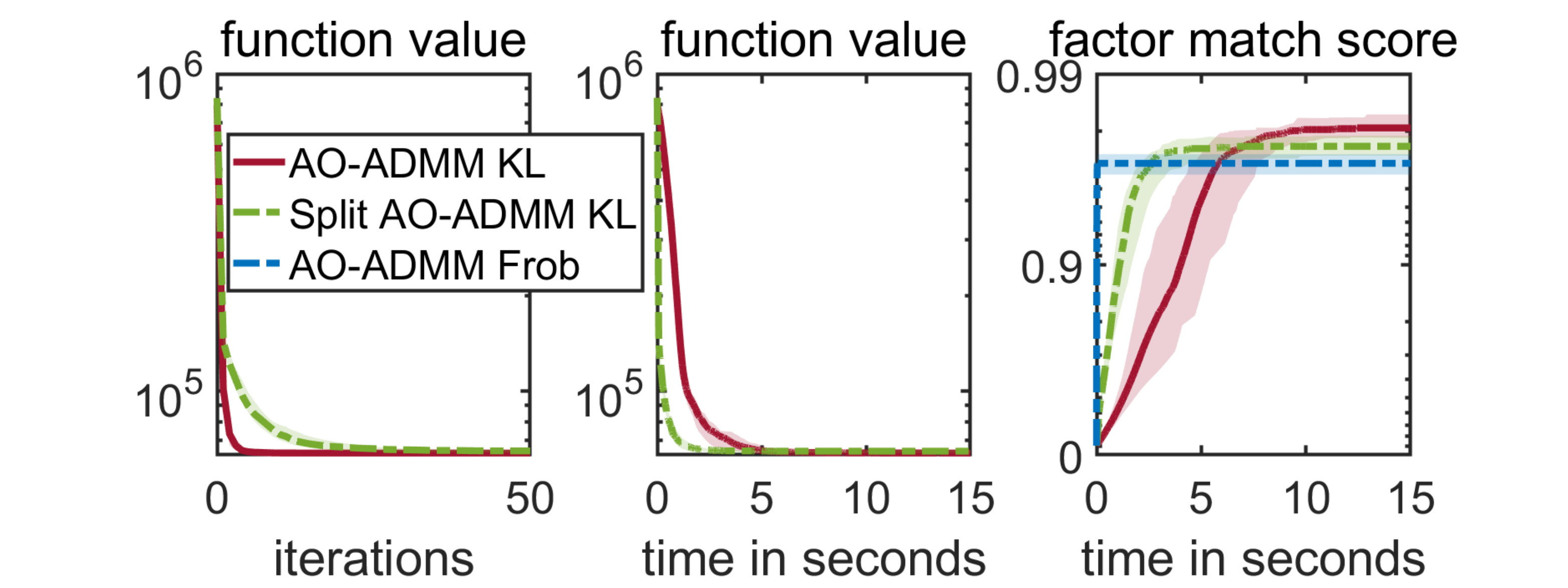}}
\caption{Experiment5: Median and quartiles of function values and FMS for  different algorithms.}
\label{fig:exp5}
\end{figure}

In this experiment, instead of adding Gaussian noise, we construct tensors containing count data $0,1,2,...$ by drawing each entry from a Poisson distribution with mean given by the CPD model of the ground-truth factor matrices. The factor matrices are drawn from a gamma distribution with shape and scale parameters equal to $1$. This ensures that factor matrices and, thus, the tensors constructed from them are non-negative. A tensor of size $40\times 50\times 60$ is coupled in the first mode with a matrix of size $40\times 100$ using coupling Case $1$. The rank of both is $R=3$. We compare the performance of our AO-ADMM implementation using KL divergence as described in Section \ref{sec:diffloss} with the variant proposed in \cite{Huang2016flexible}, which is using another split tensor variable to deal with the KL divergence. For fairer comparison and because the code is not presently accessible online, we use our own implementation.
Results based on $50$ random datasets can be seen in Fig. \ref{fig:exp5}. For each dataset, $5$ random initializations are used. Initial factor matrices are drawn from a standard uniform distribution. While in our implementation non-negativity constraints are handled by using LBFGS-B \footnote{with parameters $\texttt{m}=5$, $\texttt{maxIts}=100$, $\texttt{maxTotalIts}=5000$, $\texttt{factr}=10^{-10}/\texttt{eps}$, $\texttt{pgtol}=10^{-10}$} with lower bound $0$, in the split AO-ADMM variant, non-negativity constraints are explicitly imposed and handled via ADMM. The parameters used for the split AO-ADMM are the same as for AO-ADMM. The split tensor variable is initialized equal to the data tensor, while its dual tensor is initialized with all zeros.
The split AO-ADMM approach is computationally more efficient and thus converges faster than our approach, although it needs considerably more outer iterations. Fig. \ref{fig:exp5} shows that both algorithms achieve FMS that are above the FMS   achieved by using the squared Frobenius norm loss. However, our approach is  even more accurate than the split AO-ADMM and achieves higher factor match scores after convergence. It has no failed runs, while the split AO-ADMM failed for $3$ out of $50$ random datasets, see Table \ref{tab1:failedruns_exp5}.

\begin{table}[t!]
\caption{Failed runs experiment $5$}
\begin{center}
\vspace*{-0.5em}
\begin{tabular}{|c|c|c|c|c|}
\hline
\multirow{2}{*}{Exp.}  & \multirow{2}{*}{out of} & AO-ADMM & split AO-ADMM & AO-ADMM\\
& & KL & KL & Frobenius\\
\hline \hline
5 & all/ best & 0/0 & 19/3  &0/0\\
\hline
\end{tabular}
\label{tab1:failedruns_exp5}
\vspace*{-1.5em}
\end{center}
\end{table}

\subsection{Real Data}
\begin{figure}
\centering{\includegraphics[width=5.2in]{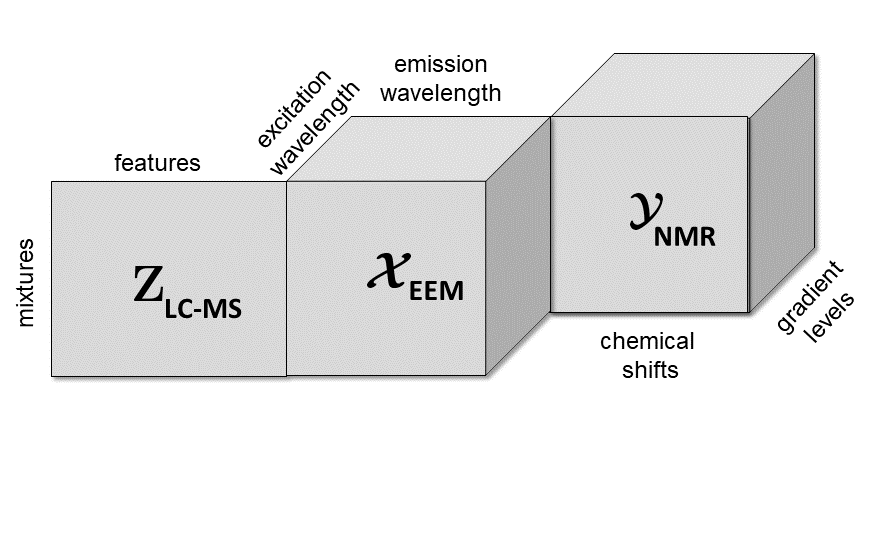}}
\caption{Datasets from different platforms, coupled in the \emph{mixtures} mode.}
\label{fig:realdata}
\end{figure}

Finally, we demonstrate an application of the proposed framework by jointly analyzing measurements of mixtures from multiple analytical platforms, and show that underlying patterns of individual chemicals in the mixtures can be accurately revealed.
Datasets consist of measurements of 28 mixtures using fluorescence spectroscopy (excitation-emission matrices (EEM)), LC-MS and diffusion NMR spectroscopy (DOSY-NMR). Mixtures were prepared using five chemicals, \textit{i.e.,} Valine-Tyrosine-Valine (Val-Tyr-Val), Tryptophan-Glycine (Trp-Gly), Phenylalanine (Phe), Maltoheptaose (Malto) and Propanol, based on a predetermined design. The fluorescence data is represented as a third-order tensor, $\tX_{EEM}$, with modes: \emph{mixtures}, \emph{excitation} and \emph{emission wavelengths}; DOSY-NMR as a third-order tensor, $\tY_{NMR}$, with modes: \emph{mixtures}, \emph{chemical shifts}, and \emph{gradient levels}, and LC-MS data as a \emph{mixtures} by \emph{features} matrix, $\mZ_{LCMS}$. Datasets are coupled in the \emph{mixtures} mode (Fig. \ref{fig:realdata}).
Three chemicals are visible in all platforms while another chemical can be detected using both NMR and LC-MS. The fifth chemical, \textit{i.e.,} Proponal, is only visible by NMR, indicating that $\tX_{EEM}$,  $\tY_{NMR}$, and $\mZ_{LCMS}$ have three, five and four components, respectively, if each component models one chemical. In total, we use a 6-component model since, in addition to the five chemicals, there is an additional component modelling the structured noise in LC-MS. See \cite{AcPaGu14, ToAcBr20} for a detailed discussion.

We jointly analyze these datasets by coupling the factor matrices in the \emph{mixtures} mode using the linear coupling given in Case 3b, as follows:
\begin{equation}\small{
\begin{aligned}
\label{eq:realdata}
&{{\min_{\left\lbrace \mC_{i,d}\right\rbrace_{\underset{d \leq D_i}{i=1,2,3}},\mDelta_1}}}
        \norm{\tX_{EEM} -\KOp{\mC_{1,d}}_{d=1}^{3} }_F^2
         +\norm{\tY_{NMR} -\KOp{\mC_{2,d}}_{d=1}^{3} }_F^2
          +  \norm{\mZ_{LCMS} -\KOp{\mC_{3,d}}_{d=1}^{2}}_F^2 \\
& \hspace{35pt} \text{s.t. }\hspace{20pt} \mC_{i,d} \geq 0, i=1,2,3, d \leq D_i\\
& \hspace{70pt}  \mC_{1,1}= \mDelta_1 \hat{\mH}_{1,1}^{\Delta}, \mC_{2,1}= \mDelta_1 \hat{\mH}_{2,1}^{\Delta}, \mC_{3,1}= \mDelta_1 \hat{\mH}_{3,1}^{\Delta}, \\
\end{aligned}}
\end{equation}
where  $\hat{\mH}_{1,1}^{\Delta} =\left[ e_1\ e_2\ e_3\right]$, $\hat{\mH}_{2,1}^{\Delta} =\left[ e_1\ e_2\ e_3\ e_4\ e_5\right]$ and $\hat{\mH}_{3,1}^{\Delta} =\left[ e_1\ e_2\ e_3\ e_4\ e_6\right]$, with $e_i$ denoting the $i$-th unit vector in $\R^6$.
Nonnegativity constraints are imposed in all modes since all factor matrices are expected to be nonnegative (\textit{e.g.,} $\mC_{1,1}$ models the relative concentrations of chemicals in the mixtures while $\mC_{2,2}$ corresponds to emission spectra).
\begin{figure*}[!t]
\centering
\subfloat[Columns of $\mC_{1,1}$ extracted from EEM.]{\includegraphics[trim=0 0 0 0, width=0.5\columnwidth]{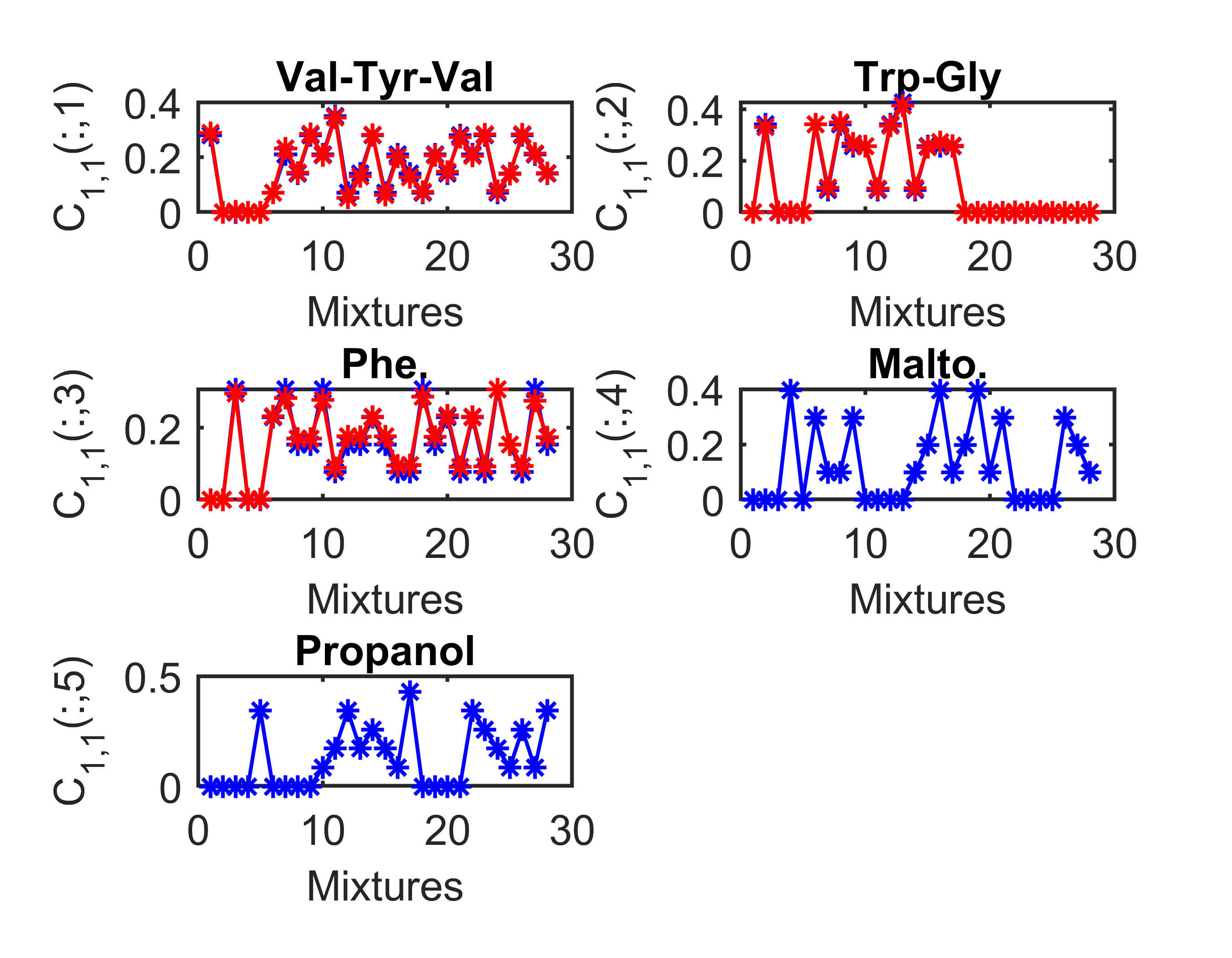}}
\label{fig:reala}
\hfil
\subfloat[Columns of $\mC_{2,1}$ extracted from NMR.]{\includegraphics[trim=0 0 0 0, width=0.5\columnwidth]{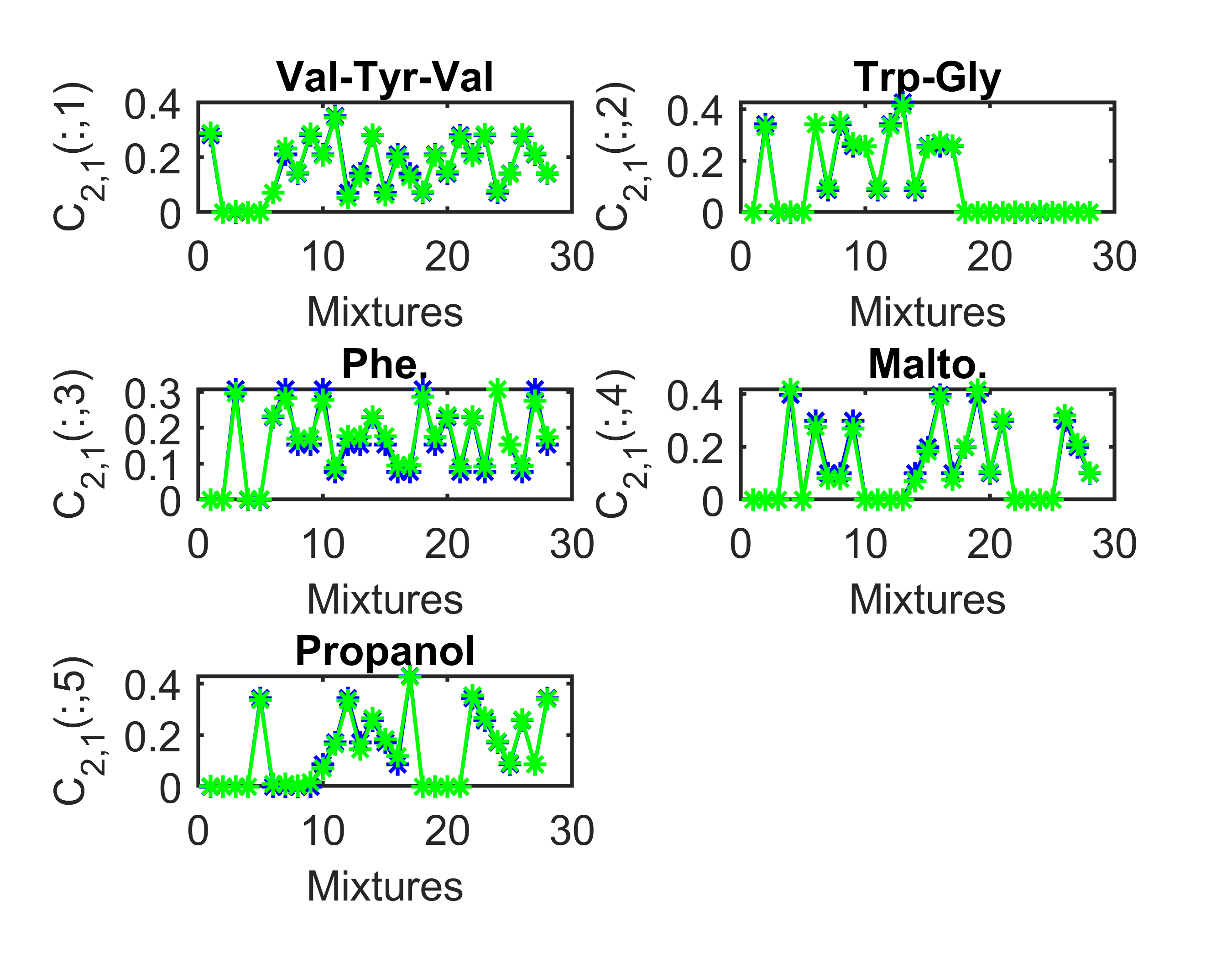}}
\label{fig:realb}
\hfil
\subfloat[Columns of $\mC_{3,1}$ extracted from LC-MS.]{\includegraphics[trim=0 0 0 0, width=0.5\columnwidth]{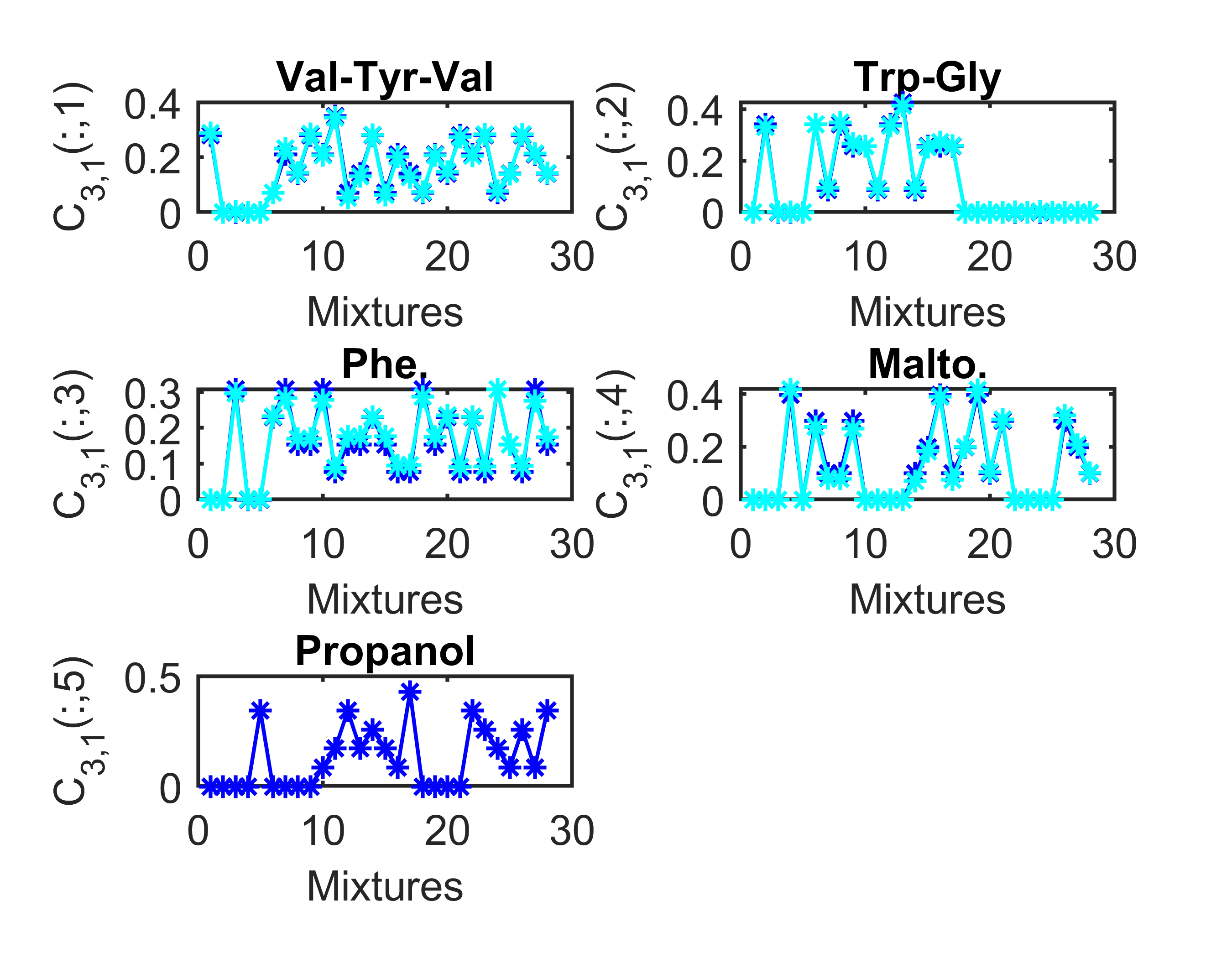}}
\label{fig:realc}
\caption{Joint analysis of measurements from multiple platforms by solving the optimization problem in (\ref{eq:realdata}). Blue lines indicate the true design of the experiment while other colors are used to plot the columns of the factor matrices extracted by the model.}
\label{fig:real}
\end{figure*}
By using Algorithm \ref{alg:AO_CMTF} to solve the optimization problem in (\ref{eq:realdata}), we can successfully extract components modelling the individual chemicals in the mixtures. Fig. \ref{fig:real} shows the factor matrices $\mC_{1,1}, \mC_{2,1},\mC_{3,1}$ coupled through the given transformation matrices in the \emph{mixtures} mode, demonstrating that individual chemicals can be separated and the underlying true design of the experiment can be recovered. While there are six components in total, each dataset captures components modelling some of the chemicals, as indicated in the transformation matrices. Similarly, factor matrices in other modes have also been extracted accurately and can be used to identify the chemical being modelled by each component (not shown). Here,  we have used multiple initializations and the one reaching the minimum function value has been reported after checking the uniqueness of the model.

\section{Discussions}\label{sec:extensions}
In this paper, we have presented a flexible algorithmic framework for regularized  and linearly coupled tensor factorizations based on ADMM which is able to handle many important loss functions, coupling structures and regularizations in a plug-and-play fashion. Numerical experiments show that our approach is computationally at least comparable, and in linearly coupled cases superior, to other state-of-the-art methods while being more flexible.

Nevertheless, there exist limitations of the current algorithmic framework.
Firstly, while the proposed algorithm can incorporate many commonly used non-parametric constraints via the proximal operator, parametric constraints can not be handled straightforwardly.
Moreover, using gradient-based numerical optimization to update the estimates of factor matrices, requires, on the one hand, the loss function $\mathcal{L}_i(\cdot,\cdot)$ to be differentiable in the second argument. This excludes, amongst others, the $\ell_1$-norm loss.
On the other hand, even for differentiable loss functions, the constraints on the entries of the reconstructed tensor that arise from the assumptions on the mean of the data distribution, may be difficult to handle, for example, for binary data following a Bernoulli distribution.
Note that in \cite{HoKoDu19}, also non-linear link functions, which map the mean of the distribution to the CPD tensor entries, are discussed which have the advantage that certain constraints can be avoided. However, this non-linear transformation changes the factor matrices irreversibly and one should be cautious when interpreting the resulting factors, especially in a coupled setting and when additional constraints are imposed on the factor matrices. For this reason, we refrain from covering non-linear link functions in this paper and in the code.

Furthermore, when coupled datasets have different statistical properties, it is desirable to use different loss functions $\mathcal{L}_i$ to fit each tensor $\tT_i$. However, how to choose the weights $w_i$ in equation \eqref{eq:genoptproblem}, that compensate for different scales and noise levels in the datasets, becomes a critical issue. The problem of optimal weights is still more or less unresolved. In \cite{SiErCeAc13}, the authors propose a probabilistic approach for estimating the dispersion parameters while estimating the factor matrices for coupled tensor factorization models with mixed divergences. Another approach is to estimate the variances in the datasets beforehand \cite{WiCeMeBe11,SoWeSm20}.

A major limitation of the current framework is that linear coupling transformations have to be known. Often, however, they are not known exactly, but might be well approximated by some parameterized function as in \cite{EyHuDe17}. We are therefore currently investigating the possibilities of learning the (parameterized) transformation matrices within the framework. This is also especially important for the automatic identification of shared and unshared components. Another possibility to reveal shared components is the ACMTF model proposed in \cite{AcPaGu14}, which supposes that all factors are identical, but only a few components have non-zero strength in each tensor $\tT_i$. This model can theoretically be obtained within the AO-ADMM framework by setting $g_{i,d}$ as the sparsity-inducing $\ell_1$ norm on the $\ell_2$ norms of columns of coupled factors, supposing other factors are normalized.

Future work includes also the extension of this framework to other tensor factorization models than the CPD. Furthermore, it can be worthwhile to consider Nesterov-type acceleration to increase efficiency of the AO-ADMM framework. Accelerations of this type have previously been applied to ALS for fitting CP decompositions\cite{MiYeSt19,AnCoHiGi20}.

%

\bibliographystyle{IEEEtran}

\end{document}